\documentclass[manuscript,screen]{acmart}

\usepackage{colortbl}
\usepackage{todonotes}
\usepackage{arydshln}
\usepackage{tablefootnote}
\usepackage[export]{adjustbox}
\usepackage{wrapfig}
\usepackage{algpseudocode} 
\usepackage[mathscr]{eucal}
\usepackage{amsmath}
\usepackage{mathtools}
\usepackage{multicol}
\usepackage{algorithm} 
\usepackage{bm}
\usepackage{hyperref}
\usepackage{booktabs}
\usepackage[applemac]{inputenc}
\usepackage{xparse}
\usepackage{xifthen}
\usepackage{mathtools}
\usepackage{multirow}
\usepackage{textcomp}
\usepackage{todonotes}
\usepackage{pifont}
\usepackage{soul}
\usepackage{color}
\usepackage{upgreek}
\usepackage{subfig}
\usepackage{graphicx}
\usepackage{geometry}
\usepackage{footnote}
\makesavenoteenv{tabular}
\usepackage{lipsum}
\usepackage{xcolor}
\definecolor{anti-flashwhite}{rgb}{0.95, 0.95, 0.96}
\renewcommand{\textsl}{\textit}
\newcommand{\comm}[1]{}
\newcommand\numberthis{\addtocounter{equation}{1}\tag{\theequation}}
\NewDocumentCommand{\vect}{ O{} O{} m }{\mathbf{#3}\ifthenelse{\isempty{#1}}{}{^{(#1)}}\ifthenelse{\isempty{#2}}{}{_{#2}}}

\NewDocumentCommand{\mat}{ O{} O{} m }{\mathbf{#3}\ifthenelse{\isempty{#1}}{}{^{(#1)}}\ifthenelse{\isempty{#2}}{}{_{#2}}}

\NewDocumentCommand{\ten}{ O{} O{} m }{\pmb{\mathscr{#3}}\ifthenelse{\isempty{#1}}{}{^{(#1)}}\ifthenelse{\isempty{#2}}{}{_{#2}}}

\makeatletter
\newcommand{\thickhline}{%
    \noalign {\ifnum 0=`}\fi \hrule height 1pt
    \futurelet \reserved@a \@xhline
}

\DeclareMathOperator*{\argmax}{arg\,max}

\AtBeginDocument{%
  }

\setcopyright{acmlicensed}
\copyrightyear{2018}
\acmYear{2018}
\acmDOI{XXXXXXX.XXXXXXX}

\acmJournal{POMACS}
\acmVolume{37}
\acmNumber{4}
\acmArticle{111}
\acmMonth{8}

\begin{document}

\title{Matrix Factorization for Inferring Associations and Missing Links}


\author{Ryan Barron}
\authornote{Authors contributed equally to this research.}
\email{barron@lanl.gov}
\orcid{1234-5678-9012}
\affiliation{%
  \institution{Theoretical Division, Los Alamos National Laboratory}
  \city{Los Alamos}
  \state{NM}
  \country{USA}
}

\author{Maksim E. Eren}
\authornotemark[1]
\email{maksim@lanl.gov}
\orcid{1234-5678-9012}
\affiliation{%
  \institution{Information Systems and Modeling, Los Alamos National Laboratory}
  \city{Los Alamos}
  \state{NM}
  \country{USA}
}

\author{Duc P. Truong}
\authornotemark[1]
\email{dptruong@lanl.gov}
\orcid{1235678-9012}
\affiliation{%
  \institution{Theoretical Division, Los Alamos National Laboratory}
  \city{Los Alamos}
  \state{NM}
  \country{USA}
}

\author{Cynthia Matuszek}
\email{cmat@umbc.gov}
\orcid{1234-5678-9012}
\affiliation{%
  \institution{Department of Computer Science and Electrical Engineering, University of Maryland, Baltimore County}
  \city{Baltimore}
  \state{MD}
  \country{USA}
}
\author{James Wendelberger}
\email{jgw@lanl.gov}
\orcid{0000-0002-7534-9130}
\affiliation{%
  \institution{Computer, Computational, and Statistical Sciences, Los Alamos National Laboratory}
  \city{Los Alamos}
  \state{NM}
  \country{USA}
}
\author{Mary F. Dorn}
\email{mfdorn@lanl.gov}
\orcid{1234-5678-9012}
\affiliation{%
  \institution{Computer, Computational, and Statistical Sciences, Los Alamos National Laboratory}
  \city{Los Alamos}
  \state{NM}
  \country{USA}
}

\author{Boian Alexandrov}
\email{boian@lanl.gov}
\orcid{1234-5678-9012}
\affiliation{%
  \institution{Theoretical Division, Los Alamos National Laboratory}
  \city{Los Alamos}
  \state{NM}
  \country{USA}
}

\renewcommand{\shortauthors}{Barron, Eren, Truong et al.}

\begin{abstract}

Missing link prediction is a method for network analysis, with applications in recommender systems, biology, social sciences, cybersecurity, information retrieval, and Artificial Intelligence (AI) reasoning in Knowledge Graphs. Missing link prediction identifies unseen but potentially existing connections in a network by analyzing the observed patterns and relationships. In proliferation detection, this supports efforts to identify and characterize attempts by state and non-state actors to acquire nuclear weapons or associated technology - a notoriously challenging but vital mission for global security. Dimensionality reduction techniques like Non-Negative Matrix Factorization (NMF) and Logistic Matrix Factorization (LMF) are effective but require selection of the matrix rank parameter, that is, of the number of hidden features, $k$, to avoid over/under-fitting. We introduce novel Weighted (WNMFk), Boolean (BNMFk), and Recommender (RNMFk) matrix factorization methods, along with ensemble variants incorporating logistic factorization, for link prediction. Our methods integrate automatic model determination for rank estimation by evaluating stability and accuracy using a modified bootstrap methodology and uncertainty quantification (UQ), assessing prediction reliability under random perturbations. Additionally, we incorporate Otsu threshold selection and k-means clustering for Boolean matrix factorization, comparing them to coordinate descent-based Boolean thresholding. Our experiments highlight the impact of rank $k$ selection, evaluate model performance under varying test-set sizes, and demonstrate the benefits of UQ for reliable predictions using abstention. We validate our methods on three synthetic datasets (Boolean and Gaussian distributed) and benchmark them against LMF and symmetric LMF (symLMF) on five real-world protein-protein interaction networks, showcasing an improved prediction performance.

\end{abstract}


\received{20 February 2007}
\received[revised]{12 March 2009}
\received[accepted]{5 June 2009}

\maketitle


\section{Introduction}
\label{sec:introduction}

Link prediction is a network analysis technique to infer missing or future connections in a network based on its current structure and patterns in the known interactions. The task of predicting these absent but potentially existing links is an important problem across various domains, including biological network analysis, where it helps uncover novel protein-protein interactions (PPI) \cite{10.1093/nar/gky1131, ppi_symlmf}, social network analysis for community detection \cite{doi:10.1073/pnas.122653799}, recommender systems for personalized suggestions \cite{xu2012alternating, 5197422, Collaborative_Filtering}, cybersecurity for anomaly detection \cite{AHMED201619, 10.1145/3519602}, and explainable reasoning over Knowledge
Graphs \cite{wang2019explainable}.

Matrix factorization techniques, such as Non-negative Matrix Factorization (NMF) and Logistic Matrix Factorization (LMF), have emerged as practical dimensionality reduction approaches for link prediction \cite{Lee1999LearningTP, ppi_symlmf}. By decomposing the network of known interactions, represented by a matrix, into lower-dimensional components, these methods aim to capture the network's latent (or hidden) structural patterns, enabling the accurate prediction of missing links based on the existing interactions. However, these models are sensitive to the choice of the rank parameter, $k$, which determines the dimensionality of the reduced space. If $k$ is selected to be too small, it results in patterns to mix (under-fitting). If $k$ is chosen to be too large, the results include noise (over-fitting), potentially reducing the predictive performance of the missing links.

To address this challenge, here we propose novel extensions of NMF-based approaches: Weighted (WNMFk), Boolean (BNMFk), and Recommender (RNMFk) matrix factorization, along with ensemble variants incorporating logistic factorization. These methods integrate automatic model determination and uncertainty quantification (UQ). Our automatic model determination heuristically identifies the optimal rank, $k$, by evaluating solution stability and accuracy \cite{nebgen2021neural}. At the same time, UQ leverages a modified bootstrap methodology to quantify prediction reliability under random perturbations of the input matrix. Furthermore, we incorporate k-means clustering \cite{Jin2010} and Otsu's method \cite{4310076} into our framework for Boolean matrix factorization. These methods are employed to determine thresholds for obtaining Boolean latent factors. A comparative analysis is performed for these thresholding approaches alongside the coordinate descent-based thresholding technique, evaluating their relative effectiveness in Boolean matrix factorization. Additionally, we compare the performance of Boolean matrix factorization with that of non-Boolean factorizations \cite{desantis2022factorization}. We assess their effectiveness in predicting the correct matrix rank for both Boolean and non-Boolean cases and their accuracy in identifying missing links.

In this paper, we evaluate the effectiveness of the proposed methods through experiments conducted on three synthetic datasets and five real-world PPI networks \cite{ppi_symlmf}. The experiments on synthetic datasets emphasize the critical role of rank selection and demonstrate that uncertainty-aware approaches can enhance model reliability by incorporating a reject option, where the model abstains from making predictions when uncertainty is high. Additionally, we analyze the impact of data sparsity by systematically increasing the test set size on the synthetic datasets. The proposed methods are benchmarked against LMF \cite{Johnson2014LogisticMF} and symLMF \cite{ppi_symlmf} on the PPI networks, showcasing improvements in prediction performance. Introduced methods are also compared against NMF with automatic model determination (NMFk) \cite{alexandrov2020patent, SmartTensors, alexandrov2013deciphering,islam2022uncovering} on the synthetic datasets. Our contributions include:
 
\begin{itemize}
    \item Introduce three new link prediction methods (WNMFk, BNMFk, and RNMFk), along with ensemble variants that incorporate logistic factorization: $\text{WNMFk}_{\text{lmf}}$, $\text{BNMFk}_{\text{lmf}}$, and $\text{RNMFk}_{\text{lmf}}$.
    
    \item Demonstrate that adding logistic factorization as an ensemble component to WNMFk, BNMFk, and RNMFk improves missing link prediction on PPI datasets.
    
    \item Adopt $k$-means clustering and Otsu-based thresholding for Boolean matrix factorization.
    
    \item Compare WNMFk, BNMFk, and RNMFk in Boolean and non-Boolean settings to evaluate their accuracy in predicting the correct matrix rank and link prediction performance.
    
    \item Highlight the importance of choosing an appropriate rank to improve link prediction on Boolean and Gaussian distributed synthetic datasets.
    
    \item Analyze the impact of data sparsity on link prediction performance.
    
    \item Introduce a UQ framework for our link prediction methods, offering a reject option for uncertain predictions. We show how this option improves overall accuracy by reducing the coverage rate-- the fraction of samples on which the model abstains from making a decision ("I do not know" option).
    
    \item Present a user-friendly Python library, Tensor Extraction of Latent Features (T-ELF), that implements the proposed methods with support for multi-processing, Graphics Processing Unit (GPU) acceleration, and High-Performance Computing (HPC) environments to handle large-scale computations~\cite{TELF}\footnote{T-ELF is available at \url{https://github.com/lanl/T-ELF}}.
\end{itemize}

The remainder of the paper is structured as follows: Section \ref{sec:relevant_work} reviews related work. Section \ref{sec:methods} introduces the background on NMF, LMF, and NMFk, followed by details on the proposed methods (WNMFk, BNMFk, and RNMFk), the LMF extension to the methods, Otsu thresholding, k-means clustering for Boolean settings, and UQ system integration. Section \ref{sec:datasets} describes the datasets. The experimental setup, including cross-validation, train/test sampling, performance metrics, and computational resources, is outlined in Section \ref{sec:experimental_setup}. Experiments on rank prediction, data sparsity, and Boolean vs. non-Boolean factorization with synthetic datasets are in Sections \ref{subsec:dog_results}, \ref{subsec:swimmer_results}, and \ref{subsec:ben_results}. Section \ref{subsec:ppi_results} presents results for LMF-extended $\text{WNMFk}_{\text{\textbf{lmf}}}$, $\text{BNMFk}_{\text{\textbf{lmf}}}$, and $\text{RNMFk}_{\text{\textbf{lmf}}}$, benchmarked against LMF and symLMF on real-world PPI datasets. Section \ref{sec:python_library} briefly introduces the public Python library implementing these methods. Appendix \ref{appendix_sec} provides additional results for a more comprehensive presentation.

\section{Relevant Work}
\label{sec:relevant_work}

Matrix factorization (MF) techniques have been widely used for predicting missing links in networks by leveraging latent structure and graph embedding methods to infer potential but previously unseen connections \cite{lee1999learning,chen2020robust,xu2016perturbation,peng2023negative,Johnson2014LogisticMF,nenova2013fca,miettinen2020recent}. In the context of NMF, early research addressed incomplete observations in user-rating data \cite{wnmfk_src}, introducing weighted NMF (WNMF) for collaborative filtering tasks. Further adaptations incorporated node attributes to tackle sparsity in link prediction, such as GJSNMF \cite{csse.2022.028841} and JWNMF \cite{2022_Tang_wnmfk}, while others focused on recommender systems using constraints \cite{HOSSEINZADEHAGHDAM2022116593} or federated learning \cite{eren2022fedsplit_recommender,Collaborative_Filtering}. Ensemble NMF approaches \cite{7987705} split large graphs into subproblems for better scalability, and other work introduced graph regularization or deep architectures \cite{10.5555/3305890.3305921,Yao2024EGDNMF}, often addressing extreme sparsity with additional complexity.

LMF \cite{Johnson2014LogisticMF} offers a probabilistic alternative for modeling binary or implicit feedback, with extensions that incorporate neighborhood structure \cite{LIU2019104960,10.1371/journal.pcbi.1004760} or symmetry constraints \cite{ppi_symlmf} to capture local and global relationships in network data. Beyond real-valued methods, Boolean or binary factorizations are useful in logical structures, employing 0/1 constraints to represent observed links. Pioneering work on Boolean matrix factorization (BMF) investigated logical and arithmetic operations \cite{binary_factorize}. In this work, we integrate Otsu's method \cite{4310076} and k-means clustering \cite{Jin2010} into Boolean factorization as thresholding mechanisms. Similarly, \cite{app12115377} employed Otsu's method and k-means clustering to segment brain diffusion imaging based on a reduced feature space obtained using NMF. While \cite{app12115377} applied these techniques to classify the features of interest in the images such as white matter, gray matter, and cerebrospinal fluid using the latent factors, our approach differs in that we integrate them directly into the factorization process to impose Boolean constraints on the latent factors. Rank selection remains a critical challenge in these various MF formulations, with recent attempts at cross-validation \cite{lee2022suitor}, bootstrap-based stability analyses \cite{cai2022rank}, or other regularization approaches \cite{doi:10.1137/1.9781611974348.81} that studied balancing model complexity and predictive accuracy.

Besides missing link prediction in conventional networks, MF has significantly impacted AI reasoning within Knowledge Graphs (KGs) \cite{wang2019explainable,cao2024knowledge}, representing relationships among entities as triple-based structures. Symbolic reasoning methods in KGs often suffer from scalability issues and incomplete knowledge, leading to the adoption of MF-based embedding techniques that uncover latent structures for knowledge base completion \cite{nickel2016review}. Decomposing adjacency matrices corresponding to KG relations allows AI systems to predict new connections, improving coverage and inferential power. These factorizations also benefit recommendation systems by fusing user-item interactions with explicit semantic links from the KG, offering more explainable recommendations \cite{wang2019explainable}. In the life sciences, large-scale biological graphs have utilized MF to hypothesize previously unknown interactions among genes, proteins, or drugs \cite{zitnik2018polypharmacy}, highlighting the versatility of factorization-based approaches for reasoning-driven discovery.

Despite the progress of MF in missing link prediction and AI reasoning, two key challenges remain. First, UQ is critical yet largely unaddressed. Most methods yield only point estimates and cannot measure how trustworthy each prediction is, although such measures are vital in high-stakes domains like healthcare and finance \cite{LIU2019104960,NEURIPS2023_54a1495b}. Second, the problem of optimal rank selection persists, with underestimation or overestimation leading to suboptimal performance \cite{lee2022suitor,8852146}. To address these gaps, our work proposes novel Weighted, Boolean, and Recommender NMF\textit{k} factorization methods, each incorporating automated rank determination and deterministic thresholding. We further enrich these methods through ensemble strategies and logistic components, and we introduce a bootstrap-based UQ mechanism that allows an abstention option for highly uncertain predictions. Combining thresholding, ensemble-driven rank selection, and confidence scoring aims to improve the reliability, interpretability, and applicability of MF-based link prediction in classic networks and knowledge-graph-driven AI systems.

\section{Methods}
\label{sec:methods}

In this section, we discuss the methodologies and techniques underlining our proposed approaches. First, we review the NMF (Section \ref{subsection:nmf}) and LMF (Section \ref{subsection:lmf}) approaches, as well as the NMFk method (Section \ref{subsection:nmfk}), which integrates an automatic model determination strategy to NMF, forming the foundational background for our introduced methods.
Sections \ref{subsection:wnmfk}, \ref{subsection:rnmfk}, and \ref{subsection:bnmfk} present the WNMFk, RNMFk, and BNMFk methods, respectively. We describe the ensemble approaches with these methods by integrating LMF in Section \ref{subsection:lmf_ensemble}. In Section \ref{subsection:boolean_perturbations}, the Boolean perturbation technique is introduced, and in Section \ref{subsection:boolean_clustering}, we discuss the Boolean clustering scheme employed in our methods when performing decomposition under Boolean settings. Additionally, Section \ref{subsection:boolean_thresholding} introduces the adoption of Otsu thresholding and k-means clustering for thresholding latent factors in Boolean decomposition, along with a description of the coordinate descent-based approach. Note that the Boolean operations discussed in Sections \ref{subsection:boolean_perturbations}, \ref{subsection:boolean_clustering}, and \ref{subsection:boolean_thresholding} are integrated to BNMFk and are utilized to enable its functionality in Boolean settings. In our experiments, Section \ref{sec:experiments}, these operations are also applied to WNMFk and RNMFk to evaluate their performance under Boolean conditions. Conversely, the Boolean settings in BNMFk can be turned off to evaluate the method's performance under non-Boolean optimization, providing a comprehensive analysis of its adaptability across different data environments. Finally, we present the integration of the UQ framework into our system in Section \ref{subsection:uq}. For ease of reference, a summary of the notations used throughout the paper is provided in Table \ref{table:notations}.

\begin{table*}[t!]
\small
\caption{Summary of the notation styles used in the paper.}
\label{table:notations}
\begin{tabular}{c|l||c|l}
\hline
\textbf{Notation}  & \textbf{Description}  & \textbf{Notation}  & \textbf{Description}  \\ \hline
$x$                & Scalar                 & $\vect{x}_i$       & $i$th element in the vector\\
$\vect{x}$         & Vector                 & $\mat{X}_{ij}$     & Entry at row $i$, column $j$\\
$\mat{X}$          & Matrix                 & $\mat{X}_{i:}$     & $i$th row\\
$\ten{X}$          & Tensor                 & $\mat{X}_{:j}$     & $j$th column\\
$\ten{X}_{::i}$    & $i$th slice (3rd dim)  & $\ten{X}_{:::i}$   & $i$th slice (4th dim)\\
$\ten{X}^{\textit{name}}$  & Superscript identifier & $*$ & Dot product\\
$\odot$           & Element-wise product    & $\otimes_B$ & Boolean matrix multiplication\\
\hline
\end{tabular}
\end{table*}

\subsection{Non-negative Matrix Factorization (NMF)}
\label{subsection:nmf}

NMF is an unsupervised learning method based on low-rank approximation that reduces dimensionality. NMF approximates a given observation matrix $\mat{X}\in \mathbb{R}_{+}^{n \times m}$ with non-negative entries, as a product of two non-negative matrices, i.e., $\mat{X} \approx \mat{W}\mat{H}$ and $\hat{\mat{X}} = \mat{W}\mat{H}$, where $\mat{W} \in \mathbb{R}_{+}^{n\times k}$, and $\mat{H} \in \mathbb{R}_{+}^{k \times m}$, and usually $k\ll m, n$. Here, $n$ is the number of features (rows), $m$ is the number of samples (columns), and the small dimension $k$ is the low-rank of the approximation. We perform this factorization via a non-convex minimization with non-negativity constraint, utilizing the multiplicative updates algorithm \cite{lee1999learning}, and  Frobenius norm as the distance metric, with an objective function:
\begin{align*}
\label{eqn:nmf_min}
    \underset{\mat{W} \in \mathbb{R}_{+}^{n\times k}, \, \mat{H} \in \mathbb{R}_{+}^{k \times m}}{\operatorname{min}} \, ||\mat{X} - \mat{W}\mat{H}||_{F}^{2}, \numberthis 
\end{align*}
which allows NMF to be treated as a Gaussian mixture model \cite{fevotte2009nonnegative}. In Equation \ref{eqn:nmf_min}, the factors $\mat{W}$ and $\mat{H}$ are the solution of the optimization problem, the latent factors, estimated via alternative updates, with update rules in Equations \ref{eqn:NMF_updates_W} and \ref{eqn:NMF_updates_H} respectively, and the performance of the minimization is evaluated by the relative reconstruction error in Equation \ref{eqn:rel_error}:

\begin{multicols}{3}
\noindent
\begin{align}
\label{eqn:NMF_updates_W}
\mat{W} =\mat{W}\odot\frac{\mat{X}\mat{H}^T}{\mat{W} \mat{H} \mat{H}^T},
\end{align}
\begin{align}
\label{eqn:NMF_updates_H}
\mat{H} =\mat{H}\odot\frac{\mat{W}^T\mat{X}}{\mat{W}^T \mat{W} \mat{H}},
\end{align}
\begin{align}
\label{eqn:rel_error}
\text{Relative Error} =\frac{||\mat{X}-\mat{W}\mat{H}||_F^2}{||\mat{X}||_F^2}.
\end{align}
\end{multicols}

\subsection{Logistic Matrix Factorization (LMF)}
\label{subsection:lmf}

 LMF extends the principles of matrix factorization to binary data, where the observed matrix $\mat{X} \in \{0,1\}^{n \times m}$ represents the presence or absence of interactions or a known link~\cite{Johnson2014LogisticMF}. Unlike NMF, which minimizes reconstruction error using the Frobenius norm, LMF incorporates the logistic regression model to capture the likelihood of binary observations. In the binary interaction matrix $\mat{X} \in \{0,1\}^{n \times m}$, each entry $\mat{X}_{ij}$ represents the presence ($X_{ij} = 1$) or absence ($X_{ij} = 0$) of a link between node $i$ and node $j$. However, in many real-world datasets, some interactions are unobserved or missing (i.e., the missing link). To account for this, binary mask matrix is defined $\mat{M} \in \{0,1\}^{n \times m}$, where $\mat{M}_{ij} = 1$ if $\mat{X}_{ij}$ is observed or a known link, and $\mat{M}_{ij} = 0$ for the missing links. For instance, in PPI networks, $\mat{X}_{ij} = 1$ might indicate a known interaction between proteins $i$ and $j$, while $\mat{X}_{ij} = 0$ represents the absence of an interaction. If the interaction status between proteins $i$ and $j$ has not been experimentally determined, $\mat{M}_{ij} = 0$ is used to indicate the missing entry at 
 $\mat{X}_{ij}$. In LMF, $\mat{X}$ is approximated as:
\begin{align*}
\label{eqn:lmf_eq}
\hat{\mat{X}} = \sigma(\mat{W}\mat{H} + \mat{b}_r + \mat{b}_c), \numberthis 
\end{align*}
where $\mat{W} \in \mathbb{R}^{n \times k}$ is the row latent feature matrix, $\mat{H} \in \mathbb{R}^{k \times m}$ is the column latent feature matrix, $\mat{b}_r \in \mathbb{R}^{n \times 1}$ is the row bias vector (broadcasted across columns), $\mat{b}_c \in \mathbb{R}^{1 \times m}$ is the column bias vector (broadcasted across rows), and the element-wise logistic sigmoid function:
\begin{align}
    \label{eqn:sigmoid}
    \sigma(x) = \frac{1}{1 + e^{-x}}.
\end{align}

Row and column biases are incorporated into the LMF model to account for systematic variations in the data. The row bias $\mat{b}_r \in \mathbb{R}^{n \times 1}$ captures row-specific effects, such as a node's general tendency to form links across all columns. Similarly, the column bias $\mat{b}_c \in \mathbb{R}^{1 \times m}$ models column-specific effects, such as a particular node's propensity to attract connections. These biases are added to the matrix reconstruction, allowing the model to better fit the data by accounting for individual node-specific tendencies. For example, in a PPI network, the row bias $\mat{b}_r$ could represent the inherent interaction likelihood of a specific protein across all other proteins. In contrast, the column bias $\mat{b}_c$ adjusts for the overall connectivity tendency of a target protein across all rows.
 
Each entry $\hat{\mat{X}}_{ij}$ in the reconstructed matrix represents the predicted probability of $\hat{\mat{X}}_{ij}$. The optimization objective in LMF minimizes the negative log-likelihood of the binary observations under the logistic model:
\begin{align*}
\underset{\mat{W}, \mat{H}, \mat{b}_r, \mat{b}_c}{\operatorname{minimize}} \, - \sum_{i=1}^n \sum_{j=1}^m \mat{M}_{ij} \left( \mat{X}_{ij} \log \hat{\mat{X}}_{ij} + (1 - \mat{X}_{ij}) \log (1 - \hat{\mat{X}}_{ij}) \right) + \lambda \left( ||\mat{W}||_F^2 + ||\mat{H}||_F^2 + ||\mat{b}_r||_2^2 + ||\mat{b}_c||_2^2 \right), \numberthis 
\end{align*}
where $\lambda$ is the regularization parameter to prevent overfitting. The optimization problem is solved using gradient-based methods. The gradients for each parameter are computed as follows:

\begin{align*}
\frac{\partial L}{\partial \mat{W}} = \mat{M} \odot (\hat{\mat{X}} - \mat{X}) \mat{H}^T + \lambda \mat{W}, \numberthis 
\end{align*}
\begin{align*}
\frac{\partial L}{\partial \mat{H}} = \mat{W}^T \left( \mat{M} \odot (\hat{\mat{X}} - \mat{X}) \right) + \lambda \mat{H}, \numberthis 
\end{align*}
\begin{align*}
\frac{\partial L}{\partial \mat{b}_r} = \sum_{j=1}^m \mat{M} \odot (\hat{\mat{X}} - \mat{X}) + \lambda \mat{b}_r, \numberthis 
\end{align*}
\begin{align*}
\frac{\partial L}{\partial \mat{b}_c} = \sum_{i=1}^n \mat{M} \odot (\hat{\mat{X}} - \mat{X}) + \lambda \mat{b}_c, \numberthis 
\end{align*}
where 
\begin{equation}
L = \frac{1}{2} \|\mat{M} \odot (\hat{\mat{X}} - \mat{X})\|_F^2 + \frac{\lambda}{2} \left( \|\mat{W}\|_F^2 + \|\mat{H}\|_F^2 + \|\mat{b}_r\|_F^2 + \|\mat{b}_c\|_F^2 \right). \numberthis 
\end{equation}

The parameters $\mat{W}$, $\mat{H}$, $\mat{b}_r$, and $\mat{b}_c$ are updated iteratively using gradient descent:
\begin{align*}
\mat{W} \leftarrow \mat{W} - \eta \frac{\partial L}{\partial \mat{W}}, \quad
\mat{H} \leftarrow \mat{H} - \eta \frac{\partial L}{\partial \mat{H}}, \quad
\mat{b}_r \leftarrow \mat{b}_r - \eta \frac{\partial L}{\partial \mat{b}_r}, \quad
\mat{b}_c \leftarrow \mat{b}_c - \eta \frac{\partial L}{\partial \mat{b}_c}, \numberthis 
\end{align*}
where $\eta$ is the learning rate. LMF is particularly well-suited for sparse, binary datasets, as it naturally handles probabilistic modeling of binary outcomes. Adding biases for rows and columns allows for improved flexibility in capturing systematic variations in data. 

\subsection{Non-negative Matrix Factorization with Automatic Model Determination (NMFk)}
\label{subsection:nmfk}

The NMF minimization requires prior knowledge of the latent dimensionality, $k$,  
and the number of latent features, which is often unknown. Choosing too small a value for $k$ leads to a poor approximation of the observables in $\mat{X}$ (a problem called under-fitting). At the same time, setting $k$ too large makes the extracted features hard to interpret, as they also capture noise in the data (a case of over-fitting). In essence, selecting $k$ is equivalent to estimating the number of model parameters, a well-known difficult problem. 
In general, the existing partial solutions to this problem are heuristic. Among these solutions is Automatic Relevance Determination (ARD) \cite{mackay1994bayesian}, which was first modified for Principal Component Analysis \cite{bishop1999bayesian} and then for NMF \cite{morup2009tuning,tan2012automatic}.
Another approach is based on the assumed stability of the NMF solution and was proposed to identify the number of stable clusters in the observational matrix $\mat{X}$ \cite{brunet2004metagenes}. A recent model selection technique, called NMFk \cite{alexandrov2020patent}, has been successfully used to decompose the most extensive collection of human cancer genomes \cite{alexandrov2013signatures}. 
NMFk integrates classical NMF-minimization with custom clustering and Silhouette statistics \cite{ROUSSEEUW198753} and combines the accuracy of the minimization and robustness/stability of the NMF solutions when a modified bootstrap procedure (i.e., generation of a random ensemble of slightly perturbed input matrices) is applied to estimate the number of latent features \cite{alexandrov2013deciphering}. Recently, NMFk was applied to many synthetic datasets with a predetermined number of latent features, and it demonstrated its superior performance of correctly estimating $k$ compared to the other known heuristics \cite{nebgen2021neural}. The exceptional performance of the NMFk method as a model selection was also demonstrated both in practice \cite{alexandrov2020repertoire} and in a large set of synthetic cancer genomes with a predetermined number of latent features  \cite{islam2022uncovering}. In addition, it was shown that NMFk performs better than spherical k-means and other methods for topic extraction \cite{vangara2020semantic}. Therefore, we use NMFk as the core factorization method with automatic model selection and adopt it to WNMFk, RNMFk, and BNMFk for heuristically estimating $k$. For completeness, we provide the pseudocode for it in Algorithm 1 and a description of it as follows:
\begin{algorithm}[htb]
    \caption{NMFk($\mat{X}$, $k^{min}$, $k^{max}$, $M$, $Sill\_{thr}=0.8$)} 
	\begin{algorithmic}[1]
	    \Require: $\mat{X} \in \mathbb{R}_{+}^{n \times m}$ , $k^{min}$, $k^{max}$ , $r$ \label{alg:nmfk}
		\For {$k$ in $k^{min}$ to $k^{max}$} \Comment{Start and end process for NMFk}
			\For{$q$ in 1 to $M$}        \Comment{Num. of Perturbations on each k}
				\State $\ten{X}_{::q}$ = Perturb($\mat{X}$) \Comment{Resampling  $\mat{X}$ to create a random ensemble}
				\State $\ten{W}_{::kq}$,$\ten{H}_{::kq}$ = NMF($\ten{X}_{::q}$,k)
				
			\EndFor
			\State  $\ten{W}^{all}$=[$\ten{W}_{::k1}$,\ldots,$\ten{W}_{::kM}$] and $\ten{H}^{all}$=[$\ten{H}_{::k1}$,\ldots,$\ten{H}_{::kM}$]
			\State $\ten{\hat{W}}$, $\ten{\hat{H}}$ = customCluster( $\ten{W}^{all}$,$\ten{H}^{all}$)
			\State $\ten{\widetilde{W}}_{::k}$ = medians( $\ten{\hat{W}}$) 
			\State $\ten{H}^{reg}_{::k}$  = NNLS($\mat{X}$,$\widetilde{\mat{W}}_{::k}$) \Comment{Column-wise regression of $\mat{H}$ with $\widetilde{\mat{W}}$ and column of $\mat{X}$}
			\State $\vect{s}_{k}$ = clusterStability($\ten{\hat{W}}$)
			\State $\vect{err_{k}}$ = reconstructErr($\mat{X}$,$\widetilde{\mat{W}}_{::k}$ , $\mat{H}^{reg}_{::k}$ ) \Comment{Column-wise reconstruction error for L-statistics}
	\EndFor

	\State $\vect{err}^{all}$=[$\vect{err}_{k^{min}}$,\ldots,$\vect{err}_{k^{max}}$] 
	\State $k^{opt}$ = PvalueAnalysis($\vect{err}^{all}$ ,$k^{min}$,$k^{max}$,$\vect{s}_{k}$,$Sill\_{thr}$) \Comment{Predicted k value using Wilcoxon}		
    \State \textbf{return} $\ten{\widetilde{W}}_{::k^{opt}}$, $\ten{H}^{reg}_{::k^{opt}}$, $k^{opt}$
	\end{algorithmic} 

\textbf{Ensure:} $k =k^{opt}$,$\ten{\widetilde{W}}_{::k^{opt}}$ $\in \mathbb{R}_{+}^{n \times k}$ ,$\ten{H}^{reg}_{::k^{opt}}$ $\in \mathbb{R}_{+}^{k \times m}$ , $\mat{X}$ = $\ten{\widetilde{W}}_{::k^{opt}}$  $\ten{H}^{reg}_{::k^{opt}}$	
\end{algorithm}
\begin{enumerate}
    \item \textit{Resampling}: Based on the observable matrix, $\mat{X}$, NMFk creates an ensemble of $M$ random matrices, $[\ten{X}_{::q}]_{q=1,...,M}$, with means equal to the original matrix $\mat{X}$. Each one of these random matrices $\ten{X}_{::q}$ is generated by perturbing the elements of $\mat{X}$ by a small uniform noise, such that: $\ten{X}_{ijq} = \mat{X}_{ij} +\delta_{ijq}$, for each $q=1,...,M$, where $\delta_{ijq}$ is the small error that is part of random distribution, different error for each $q$.
    \item \textit{NMF minimization}: We use the Frobenius norm-based multiplicative updates (MU) algorithm \cite{lee1999learning} for different numbers of latent features, $k$, in an interval $[k^{min},k^{max}]$, for each generated $M$ random matrices.
    \item \textit{Custom clustering:} For each $k\in [k^{min},k^{max}]$, NMF minimizations of the $M$ random matrices, $[\ten{X}_{::q}]_{q=1,...,M}$, results in $M$ pairs $[\ten{W}_{::kq}; \ten{H}_{::kq}]_{q=1,...,M}$. Further, NMFk clusters the set of the $M*k$ latent features, the columns of $\ten{W}_{::kq}$. The NMFk custom clustering is similar to k-means, but it holds exactly one column for each cluster from each of the $M$ NMF solutions. This constraint is needed since each NMF minimization gives exactly one solution $\ten{W}_{::kq}$ with the same number of columns, $k$. In the clustering, the cosine similarity metric measures the similarity between the columns. Here, cosine similarity determines the degree of similarity between two vectors in an inner product space.
    \item \textit{Robust $\mat{W}$ and $\mat{H}$ for each $k$:} The medians of the clusters,
    $\ten{\widetilde{W}}_{::k}$,  are the robust solution for each explored $k$. The corresponding mixing coefficients $\ten{H}^{reg}_{::k}$ are calculated by regression of $\mat{X}$ on  $\ten{\widetilde{W}}_{::k}$.
    \item \textit{Cluster stability via Silhouette statistics:} NMFk explores the stability of the obtained clusters, for each $k$, by calculating their Silhouettes \cite{ROUSSEEUW198753}. Silhouette scores quantify the cohesion and separability of the clusters. The Silhouette values range between $[-1, 1]$, where $-1$ means an unstable cluster and $+1$ means perfect stability.
    
    \item \textit{Reconstruction error:} Another metric NMFk uses is the relative reconstruction error, $R = ||\mat{X} - \ten{X}_{::k}^{rec}||/||\mat{X}||$, where $\ten{X}_{::k}^{rec} = \ten{\widetilde{W}}_{::k}*\ten{H}^{reg}_{::k}$, which measures the accuracy of the reproduction of initial data by a given solution and the number of latent features $k$. 
    \item \textit{L-statistics:} NMFk uses L-statistics \cite{vangara2021finding} to automatically estimate the number of latent features. To calculate L-statistics for each $k$, NMFk records the distributions of the column reconstruction errors, $\vect{e}_i = \Vert \mat{X}_{:j} - \ten{X}^{rec}_{:jk}\Vert/\Vert\mat{X}_{:j}\Vert$; $j=1,...,m$. L-statistics compares the distributions of column errors for different $k$ by a two-sided Wilcoxon rank-sum test to evaluate if two samples are taken from the same population \cite{Haynes2013}. 
    \item \textit{NMFk final solution:} The optimal number of latent features, $k^{opt}$, is the largest stable cluster count with low reconstruction error. The Wilcoxon rank-sum test computes the p-value for $k^{opt}$, ensuring subsequent column error distributions remain statistically similar, indicating noise fitting. L-statistics and a \emph{minimum} Silhouette threshold of $0.80$ guide selection, preventing overfitting. The corresponding $\ten{\widetilde{W}}_{::k^{opt}}$ and $\ten{H}^{reg}_{::k^{opt}}$ are the robust solutions for the low-rank factor matrices.

\end{enumerate}

\begin{figure}[t!]
    \centering
    \includegraphics[width=0.9\textwidth]{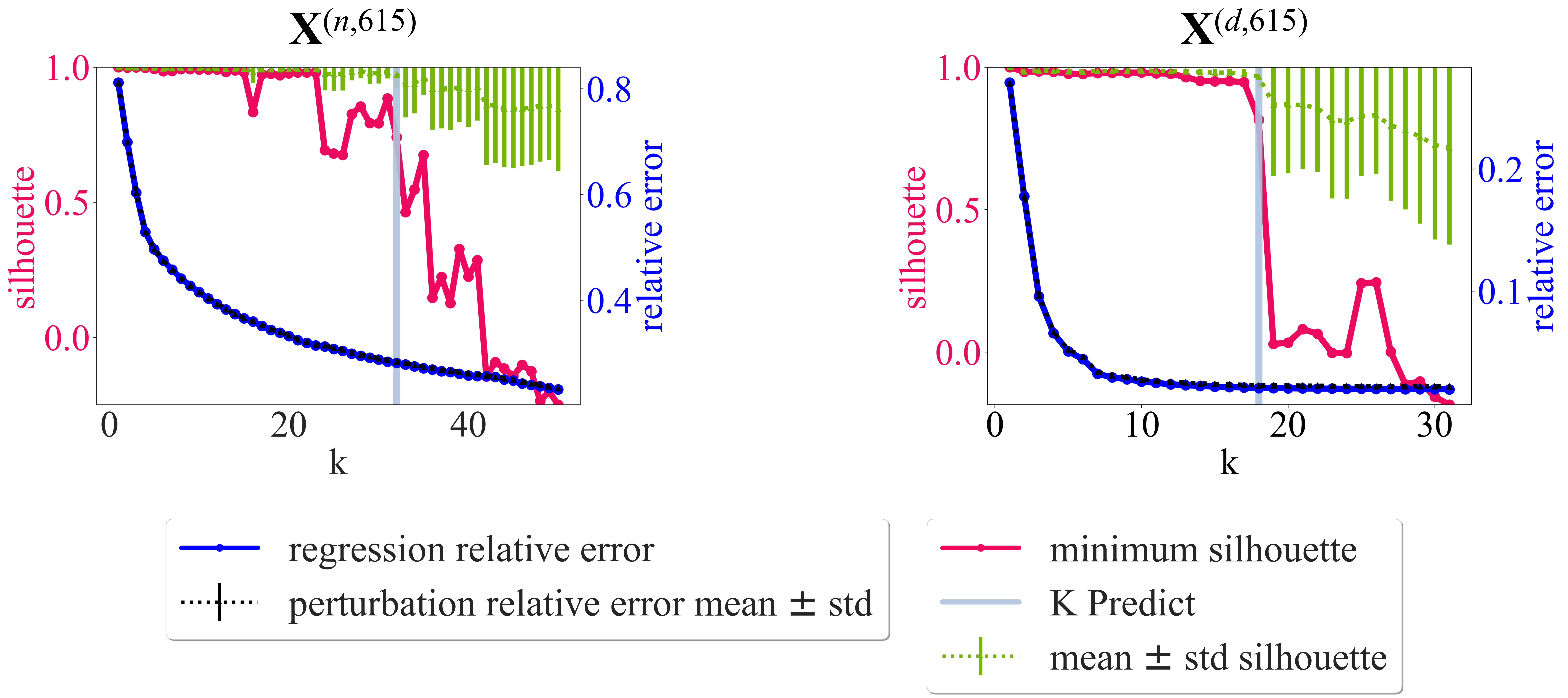}
    \caption{Sample Silhouette and relative error graphs obtained from NMFk applied to 615 malware specimens are shown \cite{10.1145/3624567}.}
    \label{fig:nmfk_sil_plots}
\end{figure}


Figure \ref{fig:nmfk_sil_plots} shows a sample Silhouette score and relative error plot from NMFk for two factorizations, demonstrating $k$ selection. NMFk estimates latent features based on two criteria: a high minimum Silhouette score and low relative reconstruction error for a stable NMF solution. Lower Silhouette scores indicate overlapping clusters, while reconstruction error decreases monotonically, with a sharp drop before stabilizing near the estimated $k$. Beyond this point, overfitting causes a sudden Silhouette decline. Therefore, $k$ with \emph{minimum} Silhouette greater than the given threshold of $0.80$ is heuristically selected as the optimal rank $k$.

\subsection{Weighted Non-Negative Matrix Factorization with Automatic Model Determination (WNMFk)}
\label{subsection:wnmfk}

While NMF, as summarized in Section \ref{subsection:nmf}, is an effective method for extracting meaningful latent features, it minimizes the distance between $\mat{X}$ and the approximation $\mat{W}\mat{H}$ as defined in Equation \ref{eqn:nmf_min}. This minimization includes zero entries in $\mat{X}$, causing the approximation $\mat{W}\mat{H}$ to be forced toward zero for those entries. However, in link prediction tasks, the network nodes interact with only a small subset of other nodes, resulting in a highly sparse matrix $\mat{X}$. Standard NMF minimization fails here because the zeros in $\mat{X}$, representing missing values, are exactly the links we aim to predict rather than the values we wish to minimize. Instead, the optimization must consider only the known entries in $\mat{X}$. Estimating the missing values in this way, as recommendations or predictions of missing links, constitutes a matrix completion problem \cite{xu2012alternating}.

WNMFk extends the NMF framework by incorporating element-wise weights during factorization. This modification enables WNMFk to account for varying confidence levels in the observed data, providing robustness in scenarios where some matrix elements are less reliable than others. Additionally, WNMFk extends standard WNMF \cite{4959890} by integrating the NMFk framework for automatic rank determination, addressing the challenge of selecting the optimal number of latent features, $k$. 

Given a matrix $\mat{X} \in \mathbb{R}_{+}^{n \times m}$ and an associated weight matrix $\mat{M} \in \{0,1\}^{n \times m}$, where $\mat{M}_{ij} > 0$ indicates that the corresponding entry $\mat{X}_{ij}$ is observed or a known link and $\mat{M}_{ij} = 0$ indicates that the entry is unobserved or a missing link, WNMFk seeks to approximate $\mat{X}$ as the product of two non-negative matrices such that $\mat{X} \approx \mat{W}\mat{H}$, where $\mat{W} \in \mathbb{R}_{+}^{n \times k}$ and $\mat{H} \in \mathbb{R}_{+}^{k \times m}$. The final prediction matrix is obtained as:
\begin{align*}
\label{eqn:wnmfk_eq}
\hat{\mat{X}} = \mat{W}\mat{H}. \numberthis 
\end{align*}
The weight matrix $\mat{M}$ prioritizes reconstructing observed entries while ignoring the missing ones. When the mask $\mat{M}$ is binary, with values $0$ and $1$, instead of continuous weights, WNMFk naturally performs missing link prediction by focusing solely on reconstructing the observed entries ($\mat{M}_{ij} = 1$) while ignoring the unobserved entries ($\mat{M}_{ij} = 0$). The objective function for WNMF is defined as:
\begin{align*}
\underset{\mat{W}, \mat{H} \geq 0}{\operatorname{minimize}} \, ||\mat{M} \odot (\mat{X} - \mat{W}\mat{H})||_F^2 + \lambda(||\mat{W}||_F^2 + ||\mat{H}||_F^2), \numberthis 
\end{align*}
where $\lambda$ is a regularization parameter to prevent overfitting. The optimization problem is solved iteratively using multiplicative update rules for $\mat{W}$ and $\mat{H}$. At each iteration, the residual matrix $\mat{R}$ is computed as $\mat{R} = \mat{X} - \mat{W}\mat{H}$. For each latent factor $k$, the updates for $\mat{W}$ and $\mat{H}$ are as follows, and $\mat{W}$ and $\mat{H}$ are constrained to remain non-negative:
\begin{align*}
W_{:, k} \leftarrow \frac{(\mat{M} \odot \mat{R}) \mat{H}_{k, :}^\top}{\mat{M}\mat{H}_{k, :}^\top \mat{H}_{k, :} + \lambda}, \numberthis 
\end{align*}
\begin{align*}
H_{k, :} \leftarrow \frac{(\mat{M} \odot \mat{R})^\top W_{:, k}}{\mat{M}^\top W_{:, k} W_{:, k} + \lambda}. \numberthis 
\end{align*}

To automatically determine the optimal rank $k$, WNMFk leverages the NMFk framework, as described in Section \ref{subsection:nmfk}. More specifically, line 4 in Algorithm 1, where $\ten{W}_{::kq}$,$\ten{H}_{::kq}$ = NMF($\ten{X}_{::q}$,k), is replaced by the factorization procedure as described in this section. WNMFk identifies the rank that balances reconstruction accuracy and stability across perturbations by applying the perturbation-based ensemble approach and clustering stability metrics. Furthermore, with WNMFk, the mask matrix $\mat{M}$ can contain non-Boolean entries, allowing continuous values to be used as weights during approximation. For instance, if specific entries in the observed matrix $\mat{X}$ are less reliable or should contribute less to the optimization process, their corresponding weights in $\mat{M}$, such as $\mat{M}_{ij}$, can be assigned lower values. For example, in a recommendation system, if a user's rating for a specific item is suspected to be influenced by noise (e.g., a randomly assigned rating rather than a genuine preference), the corresponding entry in $\mat{M}$ could be assigned a lower weight to reduce its influence on the factorization. This flexibility makes WNMFk particularly useful in scenarios with varying confidence levels across data points.

\subsection{Recommender Non-Negative Matrix Factorization with Automatic Model Determination and Biases (RNMFk)}
\label{subsection:rnmfk}

RNMFk incorporates biases to account for systematic user and item tendencies. Like WNMFk, RNMFk integrates the NMFk framework to automatically determine the optimal rank $k$, addressing the challenge of model selection. While WNMFk focuses on reconstructing observed entries in the data matrix using a weight mask, RNMFk introduces additional bias terms to improve performance in scenarios where user- and item-specific effects must be explicitly modeled. The mask $\mat{M}$ contains boolean values (0s or 1s) to signify the missing and known entries. The biases help capture variations such as individual user preferences or global item popularity, which cannot be represented solely by the latent factors $\mat{W}$ and $\mat{H}$. In this work, we extend and modify the Collaborative NMF algorithm presented in the \textit{Surprise} package \cite{Hug2020}, adopt the code to leverage vector multiplication in our Python package T-ELF, and integrate with the NMFk framework.

Given a data matrix $\mat{X} \in \mathbb{R}_{+}^{n \times m}$, where $n$ represents users or number of rows and $m$ represents items or number of columns, RNMFk aims to approximate $\mat{X}$ by factoring it into:
\begin{align*}
\label{eqn:rnmfk_eq}
\hat{\mat{X}}_{i,j} = \mat{W}_{i,:}\mat{H}_{:,j} + b_{W_i} + b_{H_j} + \mu, \numberthis 
\end{align*}
where: $\mat{W} \in \mathbb{R}_{+}^{n \times k}$ is the user latent factor matrix, $\mat{H} \in \mathbb{R}_{+}^{k \times m}$ is the item latent factor matrix, $b_{W_i}$ and $b_{H_j}$ are the biases for user $i$ and item $j$, respectively, $\mu$ is the global average rating, representing the group bias. This formulation extends the standard NMF objective by explicitly modeling biases, which account for systematic user- or item-level variations. The optimization objective in RNMFk is given by:
\begin{align*}
\underset{\mat{W}, \mat{H} \geq 0, \, b_W, \, b_H}{\operatorname{minimize}} \, ||\mat{X}_{i,j} - (\mat{W}_{i,:}\mat{H}_{:,j} + b_{W_i} + b_{H_j} + \mu)||_F^2 + \alpha||\mat{W}||_F^2 + \beta||\mat{H}||_F^2 + \gamma||b_W||_F^2 + \delta||b_H||_F^2, \numberthis 
\end{align*}
where $\alpha$, $\beta$, $\gamma$, and $\delta$ are regularization parameters for $\mat{W}$, $\mat{H}$, $b_W$, and $b_H$, respectively. The minimization is performed only over the observed entries in $\mat{X}$, making RNMFk suitable for matrix completion tasks. The parameters $\mat{W}$, $\mat{H}$, $b_W$, and $b_H$ are estimated iteratively using the following updates:
\begin{multicols}{2}
\noindent
\begin{align}
\label{eqn:bias_W_updates}
b_W \leftarrow b_W + \eta_W \sum_{j=1}^m (\mat{err}_{:,j} - \gamma b_W),
\end{align}
\begin{align}
\label{eqn:bias_H_updates}
b_H \leftarrow b_H + \eta_H \sum_{i=1}^n (\mat{err}_{i,:} - \delta b_H),
\end{align}
\end{multicols}

\begin{multicols}{2}
\noindent
\begin{align}
\label{eqn:masked_W_updates}
\mat{W} \leftarrow \mat{W} \odot \frac{\mat{X}\mat{H}^\top}{\hat{\mat{X}}\mat{H}^\top + \alpha \mat{W}},
\end{align}
\begin{align}
\label{eqn:masked_H_updates}
\mat{H} \leftarrow \mat{H} \odot \frac{\mat{W}^\top\mat{X}}{\mat{W}^\top\hat{\mat{X}} + \beta \mat{H}},
\end{align}
\end{multicols}
where $\hat{\mat{X}}_{i,j} = \mat{W}_{i,:}\mat{H}_{:,j} + b_{W_i} + b_{H_j} + \mu$ is the predicted matrix, and $\mat{err}_{i,j} = \mat{X}_{i,j} - \hat{\mat{X}}_{i,j}$. To integrate automatic model selection to RNMFk, we replace the factorization procedure of NMFk, line 4 in Algorithm 1, where $\ten{W}_{::kq}$,$\ten{H}_{::kq}$ = NMF($\ten{X}_{::q}$,k), with the factorization procedure described in this section.

\subsection{Boolean Non-Negative Matrix Factorization with Automatic Model Determination (BNMFk)}
\label{subsection:bnmfk}

 BNMFk extends the NMF framework to Boolean settings, where the data matrix $\mat{X} \in \{0,1\}^{n \times m}$ consists of binary values and integrates NMFk for automatic rank determination~\cite{truongBANMF,binary_factorize}. Unlike WNMFk and RNMFk, which use weighted and biased approaches, BNMFk applies constraints to maintain the Boolean structure in both the data and the latent factor matrices during factorization.

Given a binary matrix $\mat{X} \in \{0,1\}^{n \times m}$, BNMFk approximates $\mat{X}$ as $\mat{X} \approx \mat{W}\otimes_B\mat{H}$, where $\mat{W} \in \{0,1\}^{n \times k}$ is the Boolean row latent factor matrix, and $\mat{H} \in \{0,1\}^{k \times m}$ is the Boolean column latent factor matrix. The final prediction matrix was also obtained using the Equation \ref{eqn:wnmfk_eq}. BNMFk aims to minimize the reconstruction error while maintaining the binary constraints on $\mat{W}$ and $\mat{H}$. Unlike traditional NMF, the factorization involves thresholding operations to ensure Boolean values. We introduce the Boolean-specific clustering and thresholding operations in Sections \ref{subsection:boolean_perturbations}, \ref{subsection:boolean_clustering}, and \ref{subsection:boolean_thresholding}. In our experiments, we also present results where we turn off these Boolean settings for BNMFk and turn them on for WNMFk and RNMFk. The optimization problem for BNMFk is defined as:

\begin{align*}
\underset{\mat{W}, \mat{H} \geq 0}{\operatorname{minimize}} \, ||\mat{X} - (\mat{W} \otimes_B) \mat{H})||_F^2 + \alpha ||\mat{W}||_F^2 + \beta ||\mat{H}||_F^2, \numberthis 
\end{align*}
where $\alpha$ and $\beta$ are regularization parameters that penalize large values in $\mat{W}$ and $\mat{H}$. The Boolean constraints are enforced through adaptive thresholding during updates. BNMFk employs multiplicative updates with adaptive thresholding to ensure the Boolean structure of the latent factor matrices:
\begin{align}
\label{eqn:masked_W_updates_boolean}
\mat{W} \leftarrow \text{threshold} \left(\mat{W}\odot\frac{\mat{X} \mat{H}^\top}{\mat{W} \mat{H}^\top \mat{H} + \alpha}, \tau_\text{low}, \tau_\text{high} \right),
\end{align}
\begin{align}
\label{eqn:masked_H_updates_boolean}
\mat{H} \leftarrow \text{threshold} \left(\mat{H}\odot\frac{\mat{W}^\top \mat{X}}{\mat{W}^\top \mat{W} \mat{H} + \beta}, \tau_\text{low}, \tau_\text{high} \right),
\end{align}
where the \texttt{threshold} function, described in Section \ref{subsection:boolean_thresholding}, for  restricting $\mat{W}$ and $\mat{H}$ to Boolean values. Here, $\tau_\text{low}$ and $\tau_\text{high}$ represent the lower and upper thresholds used in the thresholding operation for binarizing the matrices $\mat{W}$ and $\mat{H}$. BNMFk integrates the NMFk framework to determine the optimal rank $k$ automatically. Like WNMFk and RNMFk, we replace the NMF procedure in NMFK, line 4 in Algorithm 1, to integrate an automatic model determination system. 

\subsection{LMF Extensions for Ensemble BNMFk, RNMFk, and WNMFk}
\label{subsection:lmf_ensemble}

We also introduce an ensemble approach for BNMFk, RNMFk, and WNMFk by integrating LMF from Section \ref{subsection:lmf} into these methods ($\text{WNMFk}_{\text{\textbf{lmf}}}$, $\text{BNMFk}_{\text{\textbf{lmf}}}$, and $\text{RNMFk}_{\text{\textbf{lmf}}}$). This extension leverages the predicted rank from the automatic model determination process in plain BNMFk, RNMFk, and WNMFk, solves for LMF with the same rank, and combines the reconstructed matrix $\hat{\mat{X}}$ from the original decomposition with the biases learned from LMF. The combined output is then passed through a sigmoid function to produce probabilistic predictions.

The ensemble approach works as follows:
\begin{enumerate}
    \item $\mat{X}$ is first factorized with one of the models introduced above BNMFk, RNMFk, or WNMFk to obtain the predicted rank $k$ and $\hat{\mat{X}}$ (Equation \ref{eqn:wnmfk_eq} for BNMFk and WNMFk, and Equation \ref{eqn:rnmfk_eq} for RNMFk) from this decomposition.
    \item The predicted rank $k$ from the automatic model determination step is used to initialize LMF, where the same $\mat{X}$ is decomposed at the predicted rank $k$ using LMF.
    \item LMF solves the optimization problem to learn row and column biases ($b_{\text{r}}$ and $b_{\text{c}}$ from Equation \ref{eqn:lmf_eq}).
    \item Then the final prediction matrix $\tilde{\mat{X}}_{\text{final}}$ is calculated as: 
    \begin{align*}
    \tilde{\mat{X}}_{\text{final}} = \sigma\left(\hat{\mat{X}} + b_{\text{row}} + b_{\text{col}}\right). \numberthis 
    \end{align*}
\end{enumerate}

This ensemble approach aims to combine the strengths of the original matrix factorization method BNMFk, RNMFk, or WNMFk with LMF, effectively combining the structural insights and adding the probabilistic modeling capabilities of LMF.

\subsection{Boolean Perturbations for Boolean Factorization}
\label{subsection:boolean_perturbations}

In the context of Boolean matrix factorization, Boolean perturbations are applied to generate an ensemble of slightly modified versions of the binary input matrix $\mat{X}$. These perturbations introduce controlled noise by flipping randomly selected entries in $\mat{X}$, ensuring diversity in the perturbed matrices while maintaining the Boolean structure. The perturbation process can be described as follows:
\begin{itemize}
    \item \textbf{Positive Noise (Additive):} A fraction of entries with value $0$ in $\mat{X}$ are randomly flipped to $1$.
    \item \textbf{Negative Noise (Subtractive):} A fraction of entries with value $1$ in $\mat{X}$ are randomly flipped to $0$.
\end{itemize}

The perturbed matrix $\mat{Y}$ is generated by applying these noise components $\mat{Y} = \text{boolean}(\mat{X}, \epsilon)$ where $\epsilon = [\epsilon_{\text{pos}}, \epsilon_{\text{neg}}]$ represents the proportion of positive and negative noise to be added. For each noise type:
\begin{itemize}
    \item $\epsilon_{\text{pos}}$: Proportion of $0$s in $\mat{X}$ flipped to $1$s.
    \item $\epsilon_{\text{neg}}$: Proportion of $1$s in $\mat{X}$ flipped to $0$s.
\end{itemize}
We replace line 3 on Algorithm 1 of NMFk, where $\ten{X}_{::q}$ = Perturb($\mat{X}$), with Boolean perturbation when applying Boolean factorization.

\subsection{Boolean Clustering for Boolean Factorization}
\label{subsection:boolean_clustering}

In the context of Boolean factorization, the clustering procedure used in the standard NMFk framework, specifically line 7 of Algorithm 1 where $\ten{\hat{W}}$, $\ten{\hat{H}}$ = customCluster( $\ten{W}^{all}$,$\ten{H}^{all}$), is replaced with a Boolean clustering approach described in this section. This adjustment ensures that the clustering step respects the Boolean nature of the data, making it more suitable for binary datasets. Boolean clustering operates on the latent factor matrices $\mat{W}$ generated from multiple perturbed instances of the input matrix $\mat{X}$. Let $\ten{W}_{\text{all}} \in \{0,1\}^{n \times k \times M}$ represent a three dimensional tensor containing the perturbed $\mat{W}$ matrices, where: $n$ is the number of rows in $\mat{W}$, $k$ is the number of latent features, $M$ is the number of perturbations, and $\mat{W}$ and $\mat{H}$ are Boolean following the thresholding techniques that will be introduced in Section \ref{subsection:boolean_thresholding}. Boolean clustering aims to iteratively align and compute the Boolean centroids for the perturbed $\mat{W}$ matrices using a distance metric tailored for Boolean data, such as Hamming distance \cite{hamming}. The Boolean clustering algorithm alternates between two main steps: \textit{distance calculation} and \textit{centroid computation}, repeated until convergence or a maximum number of iterations is reached.

\paragraph{1. Distance Calculation:}
For each perturbed $\mat{W}$, the algorithm computes the distance between the current centroids and the columns of $\mat{W}$ using a Boolean distance metric, such as Hamming distance. The columns of each perturbed $\mat{W}$ are then reordered to minimize the total distance to the centroids.

\paragraph{2. Centroid Computation:}
The centroids are updated based on the reordered $\mat{W}$ matrices. Boolean centroids are computed by aggregating the binary values of each column across all perturbations, ensuring that the centroids reflect the majority consensus of the binary data.

Let $\ten{W}_{\text{all}}$ represent the tensor of perturbed $\mat{W}$ matrices and $\mat{C}$ represent the centroids. Then the steps are:
\begin{enumerate}
    \item \textbf{Initialization:} The centroids are initialized using the first perturbed $\mat{W}$ matrix in $\ten{W}_{\text{all}}$.
    \item \textbf{Distance Calculation:} For each perturbed $\mat{W}$, compute the distance between the centroids and the columns of $\mat{W}$ and reorder the columns to minimize the distance.
    \item \textbf{Centroid Update:} Compute the Boolean centroids by aggregating the reordered columns of $\mat{W}$ across all perturbations.
    \item \textbf{Convergence Check:} If the column ordering stabilizes across all iterations, terminate the algorithm; otherwise, repeat the process.
\end{enumerate}

Boolean clustering replaces the standard custom clustering step in NMFk, enabling BNMFk (or WNMFk and RNMFk when Boolean settings are used) to effectively analyze Boolean datasets while maintaining consistency with the underlying data structure.

\subsection{Boolean Latent Factor Thresholding}
\label{subsection:boolean_thresholding}

Boolean latent factor thresholding ensures that the latent factor matrices $\mat{W}$ and $\mat{H}$ retain their Boolean structure while minimizing the reconstruction error. This section presents three thresholding techniques--Otsu's method, k-means clustering, and coordinate descent--each of which determines thresholds to binarize \(\mat{W}\) and \(\mat{H}\) while preserving a close approximation to the observed matrix \(\mat{X}\). Binarization converts continuous or multi-valued data into binary values (0s and 1s) based on a predefined threshold.

\subsubsection{Otsu's Method}
Otsu's method \cite{4310076} determines the optimal threshold for binarizing the latent factor matrices $\mat{W}$ and $\mat{H}$ by maximizing the between-class variance in their values. Specifically, for each column $\mat{W}_{:,i}$ or row $\mat{H}_{i,:}$, Otsu's method computes a threshold $t^*$ that maximizes the separability between binary classes, ensuring that $\mat{W}, \mat{H} \in \{0, 1\}$. The threshold $t^*$ is defined as:
\begin{align*}
t^* = \argmax_t \left[ \pi_0(t) \pi_1(t) \left( \mu_0(t) - \mu_1(t) \right)^2 \right], \numberthis 
\end{align*}
where $\pi_0(t)$ and $\pi_1(t)$ are the probabilities (normalized counts) of values in $\mat{W}_{:, i}$ or $\mat{H}_{i,:}$ below and above the threshold $t$, respectively, $\mu_0(t)$ and $\mu_1(t)$ are the means of the values below and above the threshold $t$, respectively. Otsu's method finds the optimal threshold by exhaustively evaluating all possible threshold values and selecting the one that maximizes the between-class variance $\sigma_B^2(t)$, defined below. For a given threshold $t$, the method partitions the data into two classes: values below $t$ and values above $t$. It then calculates the probabilities $\pi_0(t)$ and $\pi_1(t)$ and the means $\mu_0(t)$ and $\mu_1(t)$ of the two classes. The threshold that maximizes the separability of these two classes, as measured by $\sigma_B^2(t)$, is chosen as the optimal threshold. This ensures that the binary partitioning captures the most significant difference between the two groups. For each component $i$ of $\mat{W}$ or $\mat{H}$:
\begin{itemize}
    \item Compute the histogram of the values in $\mat{W}_{:,i}$ or $\mat{H}_{i,:}$.
    \item Calculate $\pi_0(t)$, $\pi_1(t)$, $\mu_0(t)$, and $\mu_1(t)$ for each potential threshold $t$.
    \item Select $t^*$ that maximizes the between-class variance:
    \begin{align*}
    \sigma_B^2(t) = \pi_0(t) \pi_1(t) \left( \mu_0(t) - \mu_1(t) \right)^2.
    \end{align*}
\end{itemize}

After determining $t^*$, binarization is applied as follows:
\begin{align*}
\mat{W}_{ij} = 
\begin{cases} 
1, & \text{if } \mat{W}_{ij} \geq t^* \\
0, & \text{if } \mat{W}_{ij} < t^*
\end{cases}
\quad ,
\mat{H}_{ij} = 
\begin{cases} 
1, & \text{if } \mat{H}_{ij} \geq t^* \\
0, & \text{if } \mat{H}_{ij} < t^*. \numberthis 
\end{cases}
\end{align*}

This approach ensures that the thresholds for $\mat{W}$ and $\mat{H}$ effectively partition the data into binary groups, maximizing the separability between classes and preserving the Boolean structure of the latent factors.

\subsubsection{K-Means Clustering}
K-means clustering \cite{pedregosa2011scikit} thresholds each component of $\mat{W}$ and $\mat{H}$ by clustering their values into two groups corresponding to 0 and 1. For a vector $\vect{z}$ (e.g., $\mat{W}_{:,i}$ or $\mat{H}_{i,:}$), k-means clustering identifies two cluster centers $c_0$ and $c_1$. The threshold is then defined as:
\begin{align*}
t^* = \frac{c_0 + c_1}{2}. \numberthis 
\end{align*}
The binarization is performed by assigning:
\begin{align*}
z_j = 
\begin{cases} 
1, & \text{if } z_j \geq t^* \\
0, & \text{if } z_j < t^*. \numberthis 
\end{cases}
\end{align*}

\subsubsection{Coordinate Descent Thresholding (search)}
Coordinate descent thresholding iteratively adjusts the thresholds for $\mat{W}$ and $\mat{H}$ to minimize the reconstruction error $\mat{X} \approx \mat{W} \odot \mat{H}$. For each component $i$, the reconstruction error is computed as:
\begin{align*}
\text{Error} = ||\mat{X} - (\mat{W}_{:,i} \mat{H}_{i,:})||_F^2. \numberthis 
\end{align*}
The thresholds $t_W[i]$ and $t_H[i]$ for $\mat{W}_{:,i}$ and $\mat{H}_{i,:}$ are optimized iteratively:
\begin{itemize}
    \item Fix $\mat{H}$ and optimize $\mat{W}_{:,i}$ by selecting $t_W[i]$ to minimize the reconstruction error.
    \item Fix $\mat{W}$ and optimize $\mat{H}_{i,:}$ by selecting $t_H[i]$ to minimize the reconstruction error.
\end{itemize}
The process repeats until the thresholds converge or the maximum number of iterations is reached. The final binarization is applied as:
\begin{align*}
\mat{W}_{:,i} = 
\begin{cases} 
1, & \text{if } \mat{W}_{:,i} \geq t_W[i] \\
0, & \text{if } \mat{W}_{:,i} < t_W[i], \numberthis 
\end{cases}
\end{align*}
and similarly for $\mat{H}_{i,:}$. In our experiments in Section \ref{sec:experiments}, we use \textit{search} as a term for this thresholding technique.

\subsection{Uncertainty Quantification (UQ)}
\label{subsection:uq}

UQ is a critical component of robust predictive modeling, providing insights into the reliability of model predictions. In the context of NMFk and its variants WNMFk, RNMFk, and BNMFk presented in this paper, UQ evaluates the stability of the reconstructed matrix $\hat{\mat{X}}$ (Equation \ref{eqn:wnmfk_eq} for BNMFk and WNMFk, and Equation \ref{eqn:rnmfk_eq} for RNMFk) across perturbations, using data augmentation to define the confidence, enabling the identification of confident and uncertain predictions. Our UQ framework leverages the idea that truly confident predictions are stable under perturbation or data augmentation \cite{10028760, Bahat2018ConfidenceFI}, which hypothesizes that stable predictions across multiple perturbations are more likely to be accurate. During the NMFk process, perturbations of the input matrix $\mat{X}$ generate an ensemble of latent factor matrices $\ten{W}_{\text{all}}$ and $\ten{H}_{\text{all}}$, corresponding to different realizations of $\mat{W}$ and $\mat{H}$. 

Under non-Boolean settings, perturbations are generated by uniformly sampling from $\mat{X}$ to create modified versions that are a controlled distance away from the original matrix $\mat{X}$. This is achieved by scaling the entries of $\mat{X}$ with random noise drawn uniformly from the range $[1 - \epsilon, 1 + \epsilon]$, where $\epsilon$ controls the magnitude of the perturbation. Then, the perturbed matrix $\mat{Y}$ is defined as:
\begin{align*}
\mat{Y} = \mat{X} \odot \left(1 - \epsilon + 2\epsilon \cdot \text{rand}(\text{shape}(\mat{X}))\right), \numberthis 
\end{align*}
where: $\text{rand}(\text{shape}(\mat{X}))$ generates random values uniformly distributed in $[0, 1]$, and $\epsilon$ is the perturbation parameter. In our experiments, we use $\epsilon=0.015$. Here, $\epsilon=0.015$ hyper-parameter is selected as it shown to give a stable region for $k$ selection \cite{nebgen2021neural}. The perturbation method described in Section \ref{subsection:boolean_perturbations} is used for Boolean settings. This approach modifies the Boolean structure of $\mat{X}$ by flipping selected entries (from $0$ to $1$ or $1$ to $0$) based on the specified noise proportions $\epsilon_{\text{pos}}$ and $\epsilon_{\text{neg}}$ (we used $(\epsilon_{\text{pos}}$, $\epsilon_{\text{neg}})=(0.015, 0.015)$ in our experiments). For each perturbation $p$, the reconstructed matrix $\hat{\mat{X}}^{(p)}$ is computed as:
\begin{align*}
\hat{\mat{X}}^{(p)} = \mat{W}^{(p)} \mat{H}^{(p)},\\
\hat{\ten{X}}_{::p} = \ten{W}_{::p}\ten{H}_{::p}, \numberthis 
\end{align*}
where $\mat{W}^{(p)} \in \mathbb{R}_{+}^{n \times k}$ and $\mat{H}^{(p)} \in \mathbb{R}_{+}^{k \times m}$ are the latent factors for perturbation $p$. For $P$ perturbations, we have $\ten{W}^{all}=[\ten{W}_{::1},\ldots, \ten{W}_{::p} , \ldots,\ten{W}_{::P}]$ and $\ten{H}^{all}=[\ten{H}_{::1},\ldots, \ten{H}_{::p} , \ldots,\ten{H}_{::P}]$, where $\ten{W}^{all}$ and $\ten{H}^{all}$ are three dimensional tensors of size $n \times k \times P$ and $k \times m \times P$. Across $P$ perturbations, the ensemble of reconstructed matrices $\{\hat{\mat{X}}^{(p)}\}_{p=1}^P$ reflects the variability in the model's predictions, where $\hat{\ten{X}}^{all}=[\hat{\ten{X}}_{::1},\ldots, \hat{\ten{X}}_{::p}, \ldots,\hat{\ten{X}}_{::P}]$, and where $\hat{\ten{X}}^{all}$ is a three dimensional tensor size of $n \times m \times P$. To quantify uncertainty, we compute the standard deviation of the reconstructed values for each entry $(i,j)$ in $\hat{\mat{X}}$:
\begin{align*}
\label{eq:UQ_matrix}
\mat{U}_{ij} = \sqrt{\frac{1}{P} \sum_{p=1}^P \left( \hat{\mat{X}}_{ij}^{(p)} - \bar{\hat{\mat{X}}}_{ij} \right)^2}, \numberthis 
\end{align*}
where: $\hat{\mat{X}}_{ij}^{(p)}$ is the $(i,j)$ entry of $\hat{\mat{X}}^{(p)}$, $\bar{\hat{\mat{X}}}_{ij} = \frac{1}{P} \sum_{p=1}^P \hat{\mat{X}}_{ij}^{(p)}$ is the mean of the reconstructed values at $(i,j)$ across all $P$ perturbations. The resulting uncertainty matrix $\mat{U}$ captures the variability of predictions at each entry of $\hat{\mat{X}}$. Low uncertainty values, or low standard deviation, indicate consistent predictions across perturbations, suggesting high confidence in the model's output at those entries. Conversely, high uncertainty values signal variability and reduced confidence in predictions. Based on the hypothesis of \textit{truly confident predictions are stable under error} \cite{10028760}, predictions with low uncertainty are more likely to be accurate and reliable.

\section{Datasets}
\label{sec:datasets}
\begin{figure}[t!]
\centering
\includegraphics[width=0.6\textwidth]{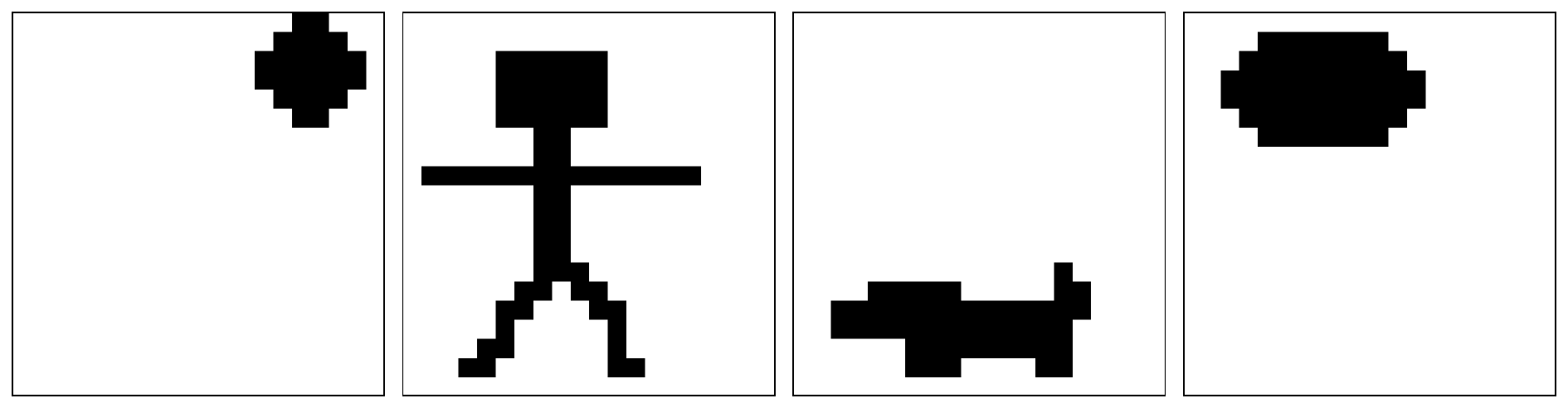} %
\caption{\textbf{Dog Dataset -} Four binary images are used as Boolean latent features to generate the synthetic data of shape $400 \times 16$.  \label{fig:dog_data}}
\end{figure}

\begin{figure}[t!]
\centering
\includegraphics[width=1.0\textwidth]{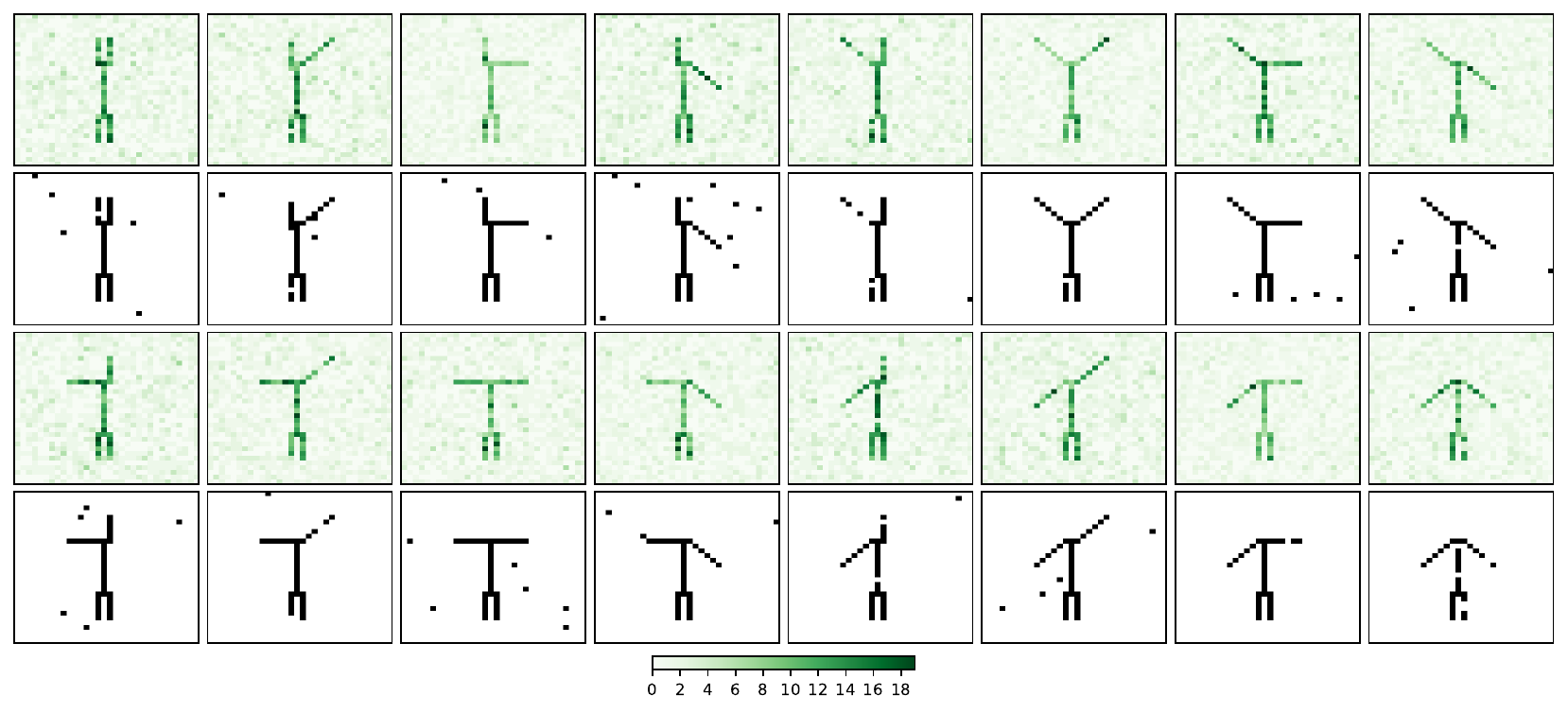} %
\caption{\textbf{Swimmer Dataset -} Dataset of 16 swimmer images. The first and third rows are the images with real-valued intensities ranging from 0 to 19. The second and fourth rows display the Boolean versions obtained after applying Otsu thresholding. For our analysis, we use the Boolean versions of the dataset, represented as a matrix of size $1024 \times 256$.  \label{fig:swimmer_data}}
\end{figure}


We evaluate our methods using three synthetic datasets (two Boolean, and one Gaussian distributed), assessing rank prediction, Boolean and non-Boolean performance, data sparsity effects, UQ, and link prediction accuracy. Additionally, we benchmark our methods against existing approaches using five real-world PPI network datasets. This section summarizes the datasets and their characteristics.

\subsection{Synthetic Datasets}

The first dataset used in our experiments, referred to as the Dog dataset \cite{binary_factorize}, is constructed by mixing four binary images. This dataset includes four distinct images--sun, person, dog, and cloud-each with dimensions $20 \times 20$ pixels. The Dog dataset is illustrated in Figure \ref{fig:dog_data}. To generate the dataset, we stack the columns of each image along the first axis, and then create 16 unique samples by combining these stacked representations using Boolean addition, considering all possible combinations of the four images. This results in a matrix $\mat{X} \in \{0, 1\}^{400 \times 16}$. Since the dataset comprises four distinct images, the true Boolean rank of the Dog dataset is $k = 4$.

The second dataset used in our experiments is the Swimmer dataset \cite{probabilistic_binary}, which contains 16 synthetic images of a "stick figure" swimming. Each image has dimensions of $32 \times 32$ pixels. Examples from this dataset are displayed in Figure \ref{fig:swimmer_data}. As shown in the first and third rows of Figure \ref{fig:swimmer_data}, the images contain "splashes" appearing as high-intensity values away from the swimmer or locations with low values of swimmer figure representing that location being under the water. We applied Otsu thresholding to convert the dataset into a Boolean format, with the resulting Boolean images displayed in rows 2 and 4 of Figure \ref{fig:swimmer_data}. The locations of the "splashes" and the places where the swimmer is under the water are visible in the Boolean version of the dataset. To construct the Boolean dataset, we stack each Boolean image along the first axis and combine them using Boolean addition into 256 different samples, resulting in a matrix $\mat{X} \in \{0, 1\}^{1024 \times 256}$. The true rank of the dataset is $k = 16$, corresponding to the 16 distinct swimmer figures.

As the third synthetic dataset, we use randomly sampled matrices $\mat{X}$ from a normal distribution, leveraging the framework introduced in \cite{nebgen2021neural}. We will call this dataset the Gaussian Dataset in our paper. We generate matrices $\mat{X} \in \mathbb{R}_{+}^{50 \times m}$, where $m \in \{50, 100, 150, 200, 250, 300, 350, 400\}$ (8 distinct shapes), corresponding to an increasing number of features. For each matrix size, we set the true rank $k \in \{2, 3, 4, 5, 6\}$ (5 distinct true rank values). Each matrix size and rank combination is randomly sampled ten times with different random seeds, resulting in $8 \times 5 \times 10 = 400$ matrices. While the two Boolean mentioned above and one non-Boolean synthetic datasets are used to assess the performance of our methods under controlled conditions, we have also utilized larger real-world datasets to evaluate further and validate our methods in practical scenarios.

\subsection{PPI Datasets}
\begin{table*}[t!]
\caption{The statistics of the five PPI datasets include the number of proteins and positive and negative interaction pairs of each PPI network when represented as a matrix.}
\label{table:ppi_dataset}
\resizebox{\textwidth}{!}{
\begin{tabular}{l|c|c|c|c|c}
\hline
\rowcolor[HTML]{D3D3D3} 
\textbf{Properties/Dataset} & \textbf{H.sapiens-extended} & \textbf{Brain} & \textbf{Disease of Metabolism} & \textbf{Liver} & \textbf{Neurodegenerative Disease} \\ \hline
Number of proteins          & 14,407                       & 11,167         & 1,036                           & 10,627          & 820                                \\
Number of positive pairs    & 157,950                      & 225,200        & 5,131                           & 218,239         & 5,881                               \\
Number of negative pairs    & 157,300                      & 223,130        & 5,123                           & 215,984         & 5,879                               \\ 
\hline  
\end{tabular}
}
\end{table*}

This paper uses five PPI network datasets from \cite{ppi_symlmf}. Specifically, the PPI networks include Brain, Disease of Metabolism, \textit{H. sapiens}-extended, Liver, and Neurodegenerative Disease. The statistics of these five PPI networks are summarized in Table \ref{table:ppi_dataset}. Our pre-processing of these datasets involves removing conflicting or inconsistent interactions. For example, cases where protein $i$ is shown to interact positively with protein $j$, but the dataset also includes a second entry indicating a negative interaction between the same proteins, are excluded. Additionally, we filter out proteins that interact with fewer than five other proteins. This step follows a traditional pre-processing technique from recommender systems, as described in \cite{Collaborative_Filtering}, serving as a pruning approach to ensure the inclusion of proteins with sufficient interactions to allow meaningful predictions.

Our experiments benchmark against the symNMF method results reported in \cite{ppi_symlmf}. At the same time, it is unclear whether \cite{ppi_symlmf} applied similar pre-processing considerations, notably removing conflicting protein interactions. Therefore, our analysis also includes a comparison with LMF (symLMF without the symmetry constraint) on the pre-processed datasets used in this paper.

\section{Experimental Setup}
\label{sec:experimental_setup}
We evaluate the results on our synthetic datasets (Dog Data, Swimmer Data, and Gaussian Data) by randomly sampling the missing links 10 times for cross-validation to ensure statistical significance. For the Boolean datasets, Dog Data and Swimmer Data, missing links are randomly sampled by stratifying on $0$s and $1$s. This enables us to test the methods' predictive capabilities on negative and positive interactions or known links. Additionally, we test the methods against data sparsity on the Dog and Gaussian datasets by systematically increasing the test-set size from 10\% to 90\% in increments of 10\%. 

Given $y$ total samples, we define the training set size as a proportion $\text{\textit{train}}_{\text{size}} \in [0.1, 0.2, \dots, 0.9]$. The number of samples in the training set is $y_{\text{train}} = y \times \text{\textit{train}}_{\text{size}}$. The test set consists of the remaining samples, with size $y_{\text{test}} = y \times (1 - \text{\textit{train}}_{\text{size}})$. For missing link prediction, we define the observation matrix $\mathbf{X}$, where known links are represented by nonzero values ($\mathbf{X}_{ij} \neq 0$), and known negative links (i.e., known absence of a connection) are represented by $\mathbf{X}_{ij} = 0$. The locations of known links are determined by the mask matrix $\mathbf{M}$, where $\mathbf{M}_{ij} = 1$ indicates an observed link in the training set, and $\mathbf{M}_{ij} = 0$ represents a missing link. To construct the training and test sets, we define the index sets:
\begin{enumerate}
\item Let $\mathcal{I}_{\text{pos}}$ be the set of indices where $\mathbf{X}_{ij} \neq 0$ (known positive links).
\item Let $\mathcal{I}_{\text{neg}}$ be the set of indices where $\mathbf{X}_{ij} = 0$ (known negative links).
\item The test set index set, $\mathcal{I}_{\text{test}}$, is chosen such that the number of indices satisfies $|\mathcal{I}_{\text{test}}| = y_{\text{test}}$. 
\end{enumerate}

The test set is constructed by stratified sampling from both known positive and known negative links:
\begin{enumerate}
    \item \textbf{Positive known links:} A subset of $\mathcal{I}_{\text{pos}}$ is randomly selected and placed in $\mathcal{I}_{\text{test}}$.
    \item \textbf{Negative known links:} A subset of $\mathcal{I}_{\text{neg}}$ is randomly selected and placed in $\mathcal{I}_{\text{test}}$.
    \item \textbf{Training set}: The remaining indices belong to the training index set $\mathcal{I}_{\text{train}} = (\mathcal{I}_{\text{pos}} \cup \mathcal{I}_{\text{neg}}) \setminus \mathcal{I}_{\text{test}}$, where $\setminus$ represents the set difference.
\end{enumerate}
The mask matrix $\mathbf{M}$ is defined such that test set locations are masked out:
\begin{equation}
\mathbf{M}_{ij} =
\begin{cases}
0, & \text{if } (i,j) \in \mathcal{I}_{\text{test}} \\
1, & \text{otherwise}.
\end{cases}
\end{equation}

The training matrix is then:

\begin{equation}
\mathbf{X}^{\text{train}} = \mathbf{X} \odot \mathbf{M},
\end{equation}

where $\odot$ represents the element-wise (Hadamard) product. The test matrix consists of the missing edges:

\begin{equation}
\mathbf{X}^{\text{test}} = \mathbf{X} \odot (1 - \mathbf{M}).
\end{equation}

A prediction matrix $\hat{\mathbf{X}}$ is computed using a model trained on $\mathbf{X}^{\text{train}}$. The performance is evaluated by comparing the predicted values at test locations:

\begin{equation}
\hat{\mathbf{X}}_{\text{test}} = \hat{\mathbf{X}} \odot (1 - \mathbf{M}),
\end{equation}

where we extract only the entries corresponding to the test set to measure prediction performance.

On the Swimmer dataset, we keep the test-set size fixed at 10\% and do not test for data sparsity, as the computational time required for this dataset is significantly longer due to the larger matrix. We set the test set size for the PPI dataset to 20\%, the same as in \cite{ppi_symlmf}. The missing links are randomly sampled 10 times to ensure robust evaluation. All results are reported with an average over the cross-validations with coverage intervals (CIs).

\subsection{Metrics}
\label{subsec:metrics}
Performance metrics include rank $k$ predictions, Root Mean Squared Error (RMSE), abstained sample fraction, RMSE on non-abstained samples, and Pearson Correlation between UQ entries and reconstruction error. We also report ROC AUC and PR AUC, including UQ-based extensions, where UQ values from $\mat{U}$ at test set points $\mathcal{I}_{\text{test}}$ serve as weights, simulating confidence-based predictions.

\subsubsection*{Rank $k$ Predictions}
We assess the accuracy of the model's automatic determination of the true rank $k$. Correct rank prediction is critical for capturing the underlying structure of the data. Our results use a Violin plot \cite{Hintze01051998} to show how the rank predictions are distributed.

\subsubsection*{Root Mean Squared Error (RMSE)}
RMSE measures the reconstruction error, quantifying how closely the reconstructed matrix $\hat{\mat{X}}$ approximates the observed matrix $\mat{X}$ at the test set points $\mathcal{I}_{\text{test}}$:
\begin{align*}
\text{RMSE} = \sqrt{\frac{1}{|\mathcal{I}_{\text{test}}|} \sum_{(i,j) \in \mathcal{I}_{\text{test}}} \left( \mat{X}_{ij} - \hat{\mat{X}}_{ij} \right)^2}. \numberthis 
\end{align*}
A lower RMSE indicates better reconstruction accuracy. We use RMSE to report the performance of predicting the missing links.

\subsubsection*{Fraction of Abstained or Rejected Samples}

This metric evaluates the proportion of predictions the model abstains from due to high uncertainty, focusing on confident predictions while disregarding less certain ones. The fraction of abstained samples is calculated as follows:
\begin{align*}
f_\text{abstain} = \frac{|\mathcal{I}_\text{abstain}|}{|\mathcal{I}_{\text{test}}|}, \numberthis
\end{align*}
where $f_\text{abstain}$ is the fraction of abstained predictions, $|\mathcal{I}_\text{abstain}|$ is the number of abstained predictions, and $|\mathcal{I}_{\text{test}}|$ is the total number of test set samples. The coverage rate can be calculated with $1-f_\text{abstain}$, referring to the fraction of non-abstained samples. The threshold for rejecting predictions in missing links is based on the uncertainty values in $\mathbf{U}$. For a given uncertainty matrix $\mathbf{U}$, the threshold for abstaining (reject-option) $\tau$ is defined as:
\begin{align*}
\tau = \frac{1}{|\mathcal{I}_{\text{train}}|} \sum_{(i,j) \in \mathcal{I}_{\text{train}}} \mathbf{U}_{ij}, \numberthis
\end{align*}
where $\mathcal{I}_{\text{train}}$ is the set of all training-set indices (known links), and $\mathbf{U}_{ij}$ is the uncertainty value at location $(i,j)$. Predictions at test-set indices $(i,j) \in \mathcal{I}_{\text{test}}$ are abstained if:
\begin{align*}
\mathbf{U}_{ij} > \tau, \numberthis
\end{align*}
where $\mathcal{I}_{\text{test}}$ represents the set of test-set indices. This ensures that the model avoids making predictions in locations with higher uncertainty than the average certainty observed in the training set (known links).

\subsubsection*{RMSE on Non-Rejected or Non-Abstained Samples}
This metric calculates the RMSE only on the predictions that are not abstained:
\begin{align*}
\text{RMSE}_{\text{Non-Abstained}} = \sqrt{\frac{1}{|\mathcal{I}_{\text{test}}'|} \sum_{(i,j) \in \mathcal{I}_{\text{test}}'} \left( \mathbf{X}_{ij} - \hat{\mathbf{X}}_{ij} \right)^2}, \numberthis
\end{align*}
where $\mathcal{I}_{\text{test}}'$ is the subset of $\mathcal{I}_{\text{test}}$ containing non-abstained predictions. This highlights the accuracy of the most confident predictions.

\subsubsection*{ROC, AUC, and PR AUC}
Receiver Operating Characteristic (ROC) AUC and Precision-Recall (PR) AUC are standard metrics for evaluating classification performance. ROC AUC measures the ability to distinguish between positive and negative classes, while PR AUC evaluates precision and recall trade-offs:
\begin{align*}
\text{ROC AUC} = \int_0^1 \text{TPR}(FPR) \, d(\text{FPR}), \quad \text{PR AUC} = \int_0^1 \text{Precision}(Recall) \, d(Recall). \numberthis
\end{align*}
Our results include their UQ-based extensions incorporating UQ values $\mathbf{U}$ as weights, simulating a scenario where confidence is assigned to predictions. The weights are calculated based on the UQ values and normalized to account for overall uncertainty. Specifically, the weight for each prediction is defined as:
\begin{align*}
w_{ij} = \frac{\mathbf{U}_{ij}}{1 + \text{median}(\mathbf{U}_{kl} \text{ for } (k,l) \in \mathcal{I}_{\text{train}})}. \numberthis
\end{align*}
The normalized weight is given by:
\begin{align*}
w_{ij}^{\text{norm}} = \frac{1}{1 + w_{ij}}. \numberthis
\end{align*}

Using these normalized weights, the UQ-based ROC AUC and PR AUC metrics are computed as:
\begin{align*}
\text{UQ - ROC AUC} = \frac{\sum_{(i,j) \in \mathcal{I}_{\text{test}}} w_{ij}^{\text{norm}} \cdot \text{TPR}(\text{FPR})}{\sum_{(i,j) \in \mathcal{I}_{\text{test}}} w_{ij}^{\text{norm}}}, \numberthis
\end{align*}
\begin{align*}
\text{UQ - PR AUC} = \frac{\sum_{(i,j) \in \mathcal{I}_{\text{test}}} w_{ij}^{\text{norm}} \cdot \text{Precision}(\text{Recall})}{\sum_{(i,j) \in \mathcal{I}_{\text{test}}} w_{ij}^{\text{norm}}}. \numberthis
\end{align*}

We incorporate these weights into the ROC AUC and PR AUC calculations by utilizing the \texttt{sample\_weight} parameter provided in Scikit-learn \cite{pedregosa2011scikit}. This allows us to account for the uncertainty-based weights during the evaluation, ensuring that higher-confidence predictions contribute more significantly to the metrics. In contrast, less confident predictions have a reduced impact, reflecting the effect of uncertainty on classification performance.

\subsection{System Configuration}
We ran the experiments on an HPC cluster named Dracarys, 
located at the Los Alamos National Laboratory (LANL). 
Dracarys 
uses the AMD EPYC 9454 48-core processor at a clock speed of 3.81GHz. There are 192 virtual processors and a total physical RAM of 1.97 TeraBytes (TBs). The system also comprises 8 NVIDIA H100 GPUs with VRAM memory of 82 GigaBytes (GBs) each.

\section{Experiments}
\label{sec:experiments}

This section presents our methods' performance, starting with synthetic datasets under Boolean and non-Boolean settings. We evaluate rank prediction, link prediction, performance under increasing sparsity, and UQ utility in controlled settings with pre-determined ranks. Next, we showcase results on PPI networks, demonstrating real-world effectiveness. Additional synthetic dataset results are reported in Appendix \ref{appendix_sec}.

\subsection{Dog Data}
\label{subsec:dog_results}

\begin{figure}[t!]
\centering
\includegraphics[width=1.0\textwidth]{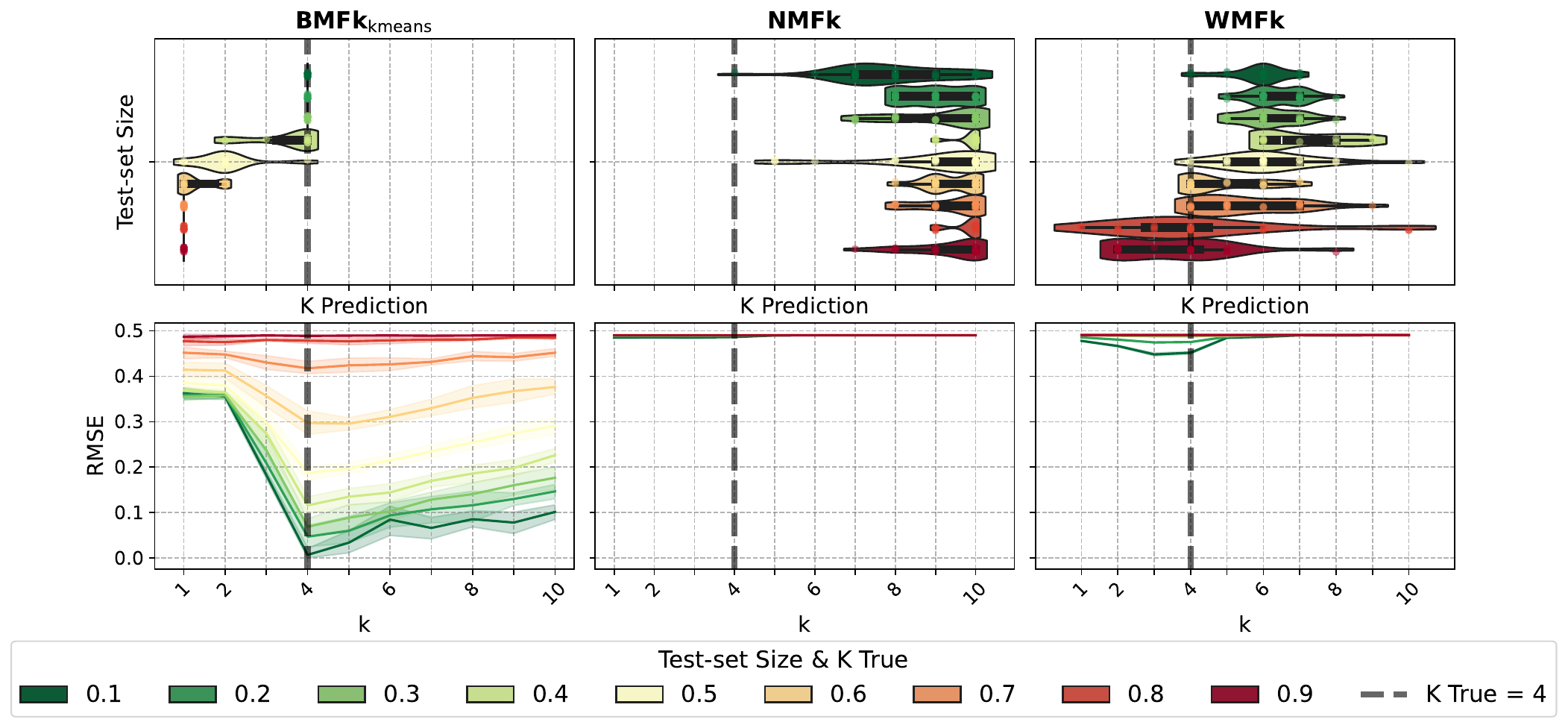} %
\caption{Results for Dog Data across methods including $\text{BNMFk}_{\text{\textbf{kmeans}}}$, NMFk, and WNMFk. Boolean thresholding is not used for NMFk and WNMFk. The first row presents violin plots visualizing the rank $k$ predictions at different test-set size levels. The second row displays RMSE scores for the test set, demonstrating the missing link prediction performance. The results are reported for each rank $k$ on the x-axis, with the dark/dashed vertical line across columns being the true rank $k=4$.
 \label{fig:dog_data_results}}
\end{figure}

\begin{figure}[t!]
\centering
\includegraphics[width=1.0\textwidth]{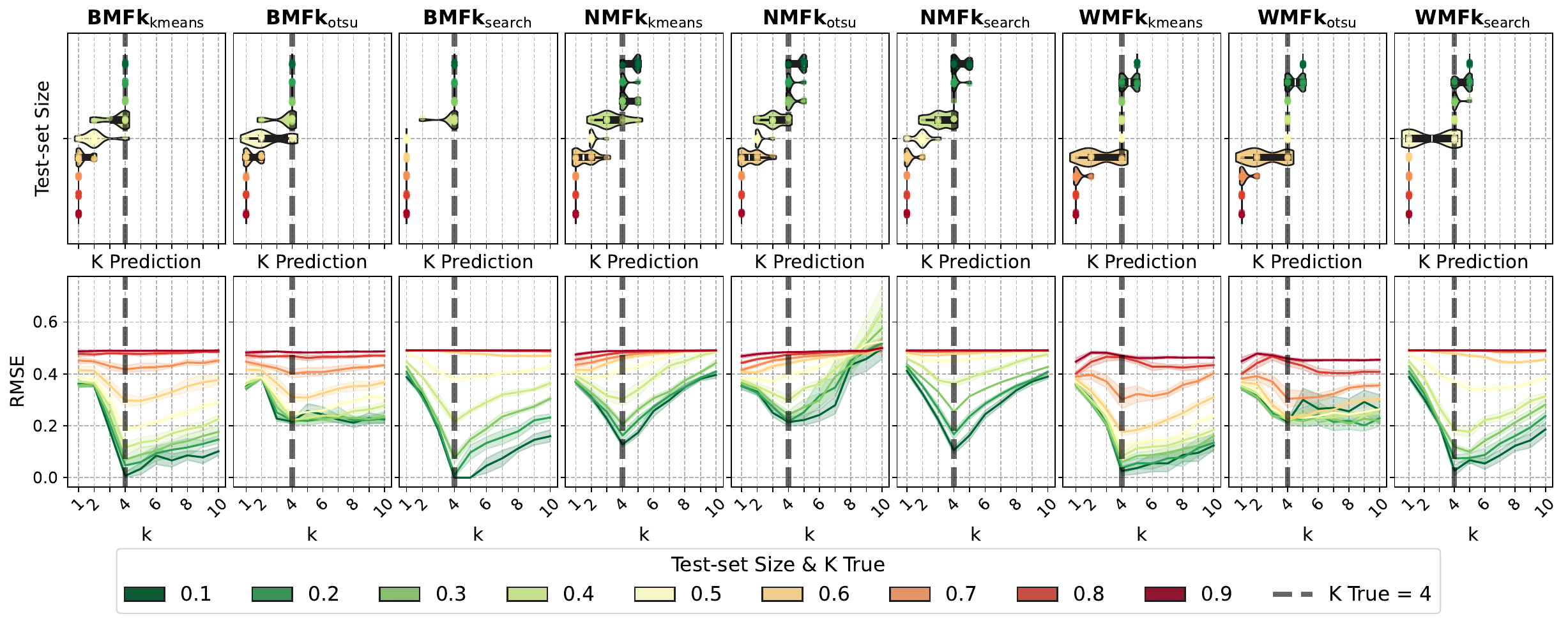} %
\caption{Results for Dog Data across methods, including BNMFk, NMFk, and WNMFk, evaluated under different Boolean thresholding techniques. The Boolean thresholding techniques are denoted with the subscripts of \textbf{kmeans}, \textbf{otsu}, and \textbf{search} (coordinate descent). The first row presents violin plots visualizing the rank $k$ predictions at different test-set sizes. The second row displays RMSE scores for the test set, demonstrating the missing link prediction performance. The results are reported for each rank $k$ on the x-axis, with the dark and dashed vertical line across each column is the true rank $k=4$.
 \label{fig:dog_data_results_booleans}}
\end{figure}

\begin{figure}[t!]
\centering
\includegraphics[width=1.0\textwidth]{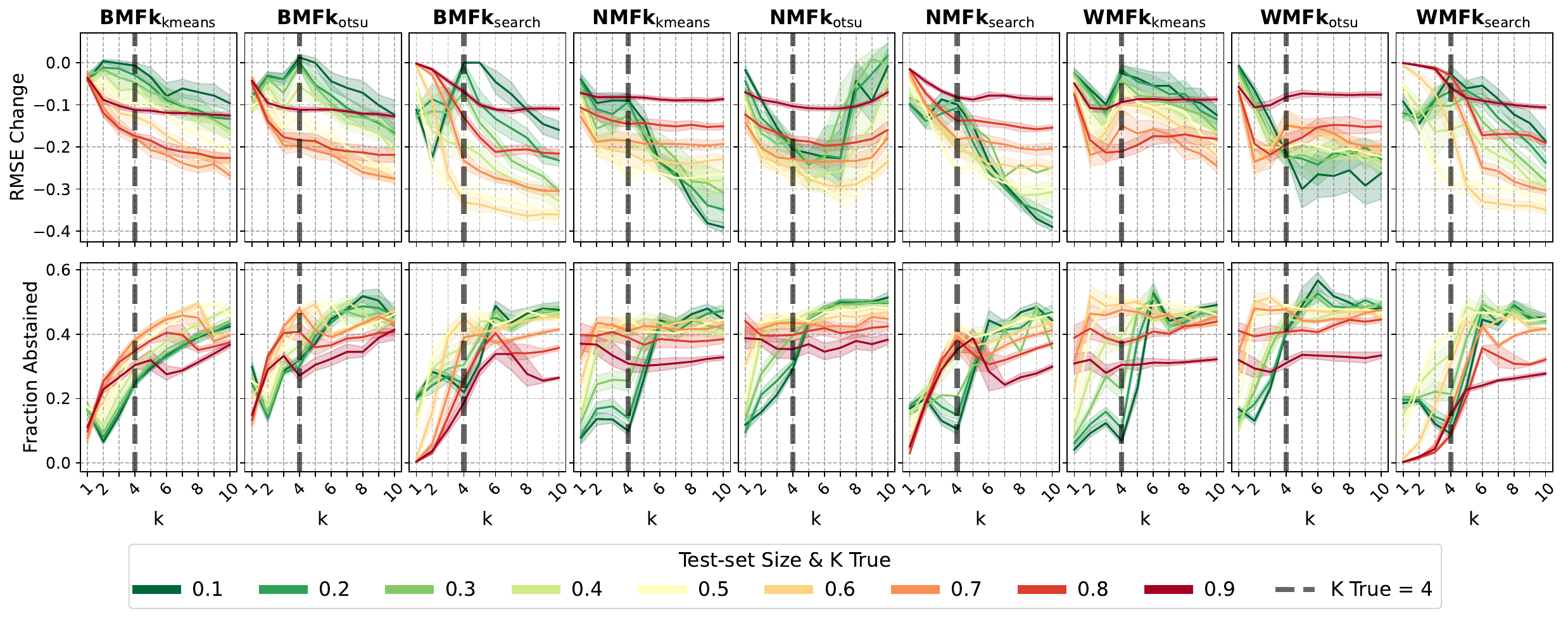} %
\caption{Results for Dog Data across methods, including BNMFk, NMFk, and WNMFk, evaluated under different Boolean thresholding techniques. The Boolean thresholding techniques are denoted with the subscripts of \textbf{kmeans}, \textbf{otsu}, and \textbf{search} (coordinate descent). The first row presents the change in RMSE after making predictions on the non-abstained samples, where a negative change indicates improved performance (lower RMSE) for the given fraction of abstained samples. The second row illustrates the fraction of abstained predictions, representing the proportion of cases where the model opted not to make a prediction due to uncertainty. The results are reported for each rank $k$ on the x-axis, with the dark and dashed vertical line across each column indicating the true rank $k=4$.
 \label{fig:dog_data_results_booleans_reject_option}}
\end{figure}

Our results on the Dog Data are presented in Figures \ref{fig:dog_data_results}, \ref{fig:dog_data_results_booleans}, and \ref{fig:dog_data_results_booleans_reject_option}, where each column displays the performance of our methods and methods under different settings. For completeness, we have a more comprehensive version of these results in Appendix \ref{appendix-dog} in Figure \ref{fig:dog_data_results_appendix}.

Figure \ref{fig:dog_data_results} compares $\text{BNMFk}_{\text{\textbf{kmeans}}}$, NMFk, and WNMFk, where the k-means subscript denotes Boolean thresholding via k-means clustering (Section \ref{subsection:boolean_thresholding}). In the first row, $\text{BNMFk}_{\text{\textbf{kmeans}}}$ predicts the true rank $k=4$ more consistently than NMFk and WNMFk, as expected for a Boolean decomposition method on Boolean Dog Data. However, for test-set sizes above 0.4, $\text{BNMFk}_{\text{\textbf{kmeans}}}$ shifts toward $k=1$, while NMFk and WNMFk frequently predict ranks above $k=4$, capturing non-Boolean structures. Both $\text{BNMFk}_{\text{\textbf{kmeans}}}$ and WNMFk predict lower ranks as sparsity increases. The second row shows RMSE scores, where lower values indicate better link prediction. $\text{BNMFk}_{\text{\textbf{kmeans}}}$ achieves the lowest RMSE at smaller test sets, with RMSE increasing as test-set size grows, limiting pattern learning. RMSE is minimized near $k=4$, emphasizing the importance of correct rank selection.


Figure \ref{fig:dog_data_results_booleans} expands on these results, comparing Boolean thresholding techniques for BNMFk, NMFk, and WNMFk (Section \ref{subsection:boolean_thresholding}). Each column represents a method using a different thresholding approach, denoted by subscripts \textbf{kmeans}, \textbf{otsu}, and \textbf{search}, with \textbf{search} referring to coordinate descent-based thresholding. For instance, $\text{BNMFk}_{\text{\textbf{otsu}}}$ applies Otsu thresholding. The first row presents rank prediction violin plots, while the second row reports RMSE scores. Compared to Figure \ref{fig:dog_data_results}, Boolean thresholding helps NMFk and WNMFk predict $k=4$ more frequently, though they still tend to overestimate ranks at smaller test sizes, whereas BNMFk predicts $k=4$ or lower. RMSE results confirm Boolean settings improve link prediction, especially for $\text{WNMFk}_{\text{\textbf{kmeans}}}$ and $\text{WNMFk}_{\text{\textbf{search}}}$. BNMFk remains the best for rank prediction and lower RMSE, except with Otsu thresholding, which yields higher RMSE values. Across all methods, a trend of improved RMSE scores at or around the true rank can be observed in this figure, further highlighting the importance of rank selection.


Figure \ref{fig:dog_data_results_booleans_reject_option} highlights UQ's role in improving predictions. The first row shows the RMSE change after applying UQ, where negative values indicate improved performance (reduced RMSE value) by abstaining from uncertain predictions. The second row presents the fraction of abstained predictions at each test-set size, with higher fractions indicating lower coverage. UQ effectively filters uncertain predictions, lowering RMSE in several cases compared to Figure \ref{fig:dog_data_results_booleans} (without UQ). Methods like $\text{BNMFk}_{\text{\textbf{search}}}$ and $\text{WNMFk}_{\text{\textbf{kmeans}}}$ show notable RMSE reduction. Abstention rates generally increase with test-set size, reflecting greater uncertainty in sparse data, until around a test-set size of 0.8. Beyond this point, the training set becomes too small, resulting in no meaningful abstentions. However, at the true rank $k=4$, abstention rates decline for several methods, which combined with observations in Figure \ref{fig:dog_data_results_booleans} showing improved performance at true rank, indicate higher confidence in predictions at the correct rank.


\begin{figure}[t!]
\centering
\includegraphics[width=1.0\textwidth]{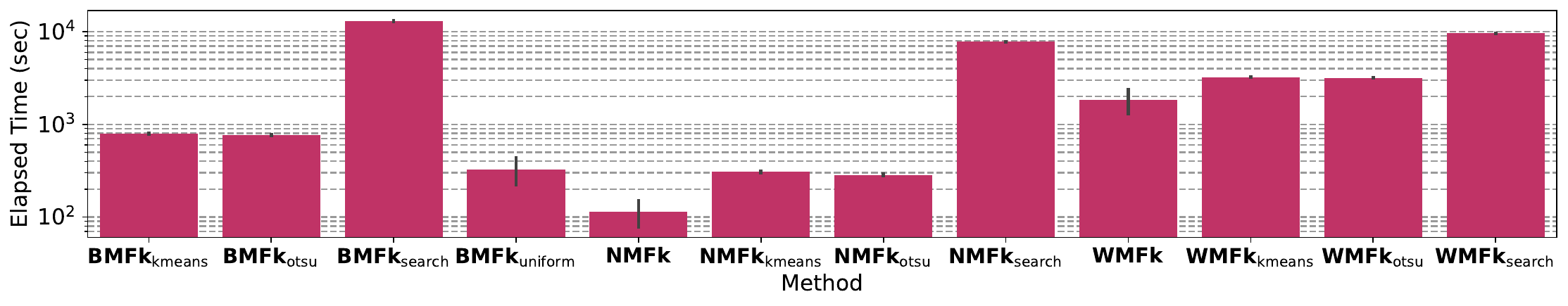} %
\caption{Computation times for each method are shown in the log scale on the Dog Data. The results indicate that methods leveraging coordinate descent (search) for Boolean thresholding require significantly longer computation times than other approaches.
\label{fig:dog_data_compute_time}}
\end{figure}

Figure \ref{fig:dog_data_compute_time} shows the computational time for each method in seconds. While Figure \ref{fig:dog_data_results} demonstrated that coordinate descent-based Boolean thresholding yields better results, its improvement comes at significantly higher computational time. As shown, methods using the search setting have substantially longer run times. Given this, we exclude the search method from subsequent experiments.

\subsection{Swimmer Data}
\label{subsec:swimmer_results} 

\begin{figure}[t!]
\centering
\includegraphics[width=1.0\textwidth]{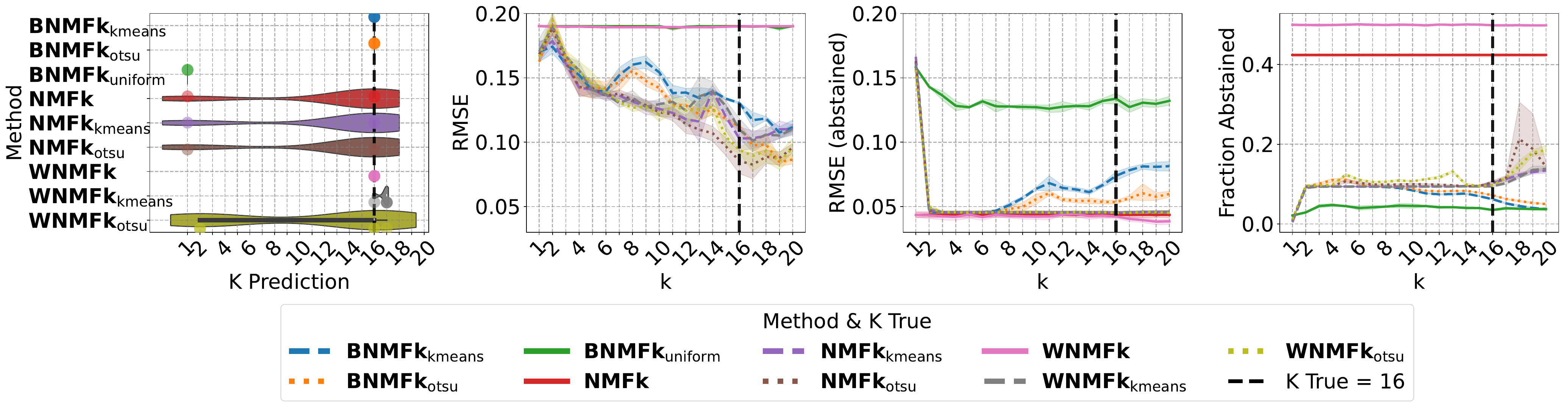} %
\caption{The performance of each method on the Swimmer Data is shown. The first column features a violin plot displaying the distribution of $k$ predictions for each method, with the y-axis representing the methods and the x-axis representing the predicted $k$ values. The second through final columns present results at each $k$ decomposition value, with the x-axis corresponding to the rank $k$. The dark dashed vertical lines indicate this dataset's true rank of $k=16$. The Boolean thresholding techniques are denoted with the sub-scripts of \textbf{kmeans}, \textbf{otsu}, and \textbf{search} (coordinate descent), while \textbf{uniform} sub-script refers to running BNMFk without Boolean thresholding on the latent factors. NMFk and WNMFk without sub-scripts are the results of not using Boolean techniques with these methods.
 \label{fig:swimmer_data_results}}
\end{figure}

We analyze the Swimmer dataset, with results in Figure \ref{fig:swimmer_data_results}. The test-set size is fixed at 0.1 (10\%) without sparsity analysis. Figure \ref{fig:swimmer_data_results} shows $k$-predictions (first column), RMSE scores (second), RMSE with abstention (third), and the fraction of abstained samples (fourth). The true rank $k=16$ is marked by a dark dashed vertical line, with the x-axis representing $k$ values. Each method is color-coded, while line styles distinguish Boolean settings in columns 2-4. Figure \ref{fig:swimmer_data_results} shows that BNMFk with Boolean settings consistently identified $k=16$, as did WNMFk without Boolean settings. WNMFk with k-means thresholding predicted both $k=16$ and $k=17$, while $\text{BNMFk}_{\text{\textbf{uniform}}}$ (without Boolean settings) always predicted $k=1$. Other methods produced $k$ values between $k=1$ and $k=17$. Some non-Boolean methods predicted $k=16$ because the Swimmer dataset's non-negative and Boolean ranks are equal.

In the second column of Figure \ref{fig:swimmer_data_results}, $\text{NMFk}_{\text{\textbf{kmeans}}}$, $\text{WNMFk}_{\text{\textbf{otsu}}}$, $\text{WNMFk}_{\text{\textbf{kmeans}}}$, and $\text{BNMFk}_{\text{\textbf{kmeans}}}$ improve as $k$ nears 16, after which RMSE increases. Notably, $\text{BNMFk}_{\text{\textbf{kmeans}}}$ continues improving until $k=18$ before RMSE rises, mirroring Dog Data trends. Non-Boolean decompositions maintain higher RMSE (lower performance) across all $k$, confirming Boolean methods perform better on Boolean data. The third column shows UQ significantly improves RMSE, even for non-Boolean methods, though less so for $\text{BNMFk}_{\text{\textbf{uniform}}}$. The RMSE remains stable across ranks as models abstain from uncertain predictions. However, as a trade-off, non-Boolean methods like WNMFk and NMFk exhibit a higher reduction in coverage, shown by a greater fraction of abstained samples in the last column.

\subsection{Gaussian Data}
\label{subsec:ben_results}

\begin{figure}[t!]
\centering
\includegraphics[width=0.9\textwidth]{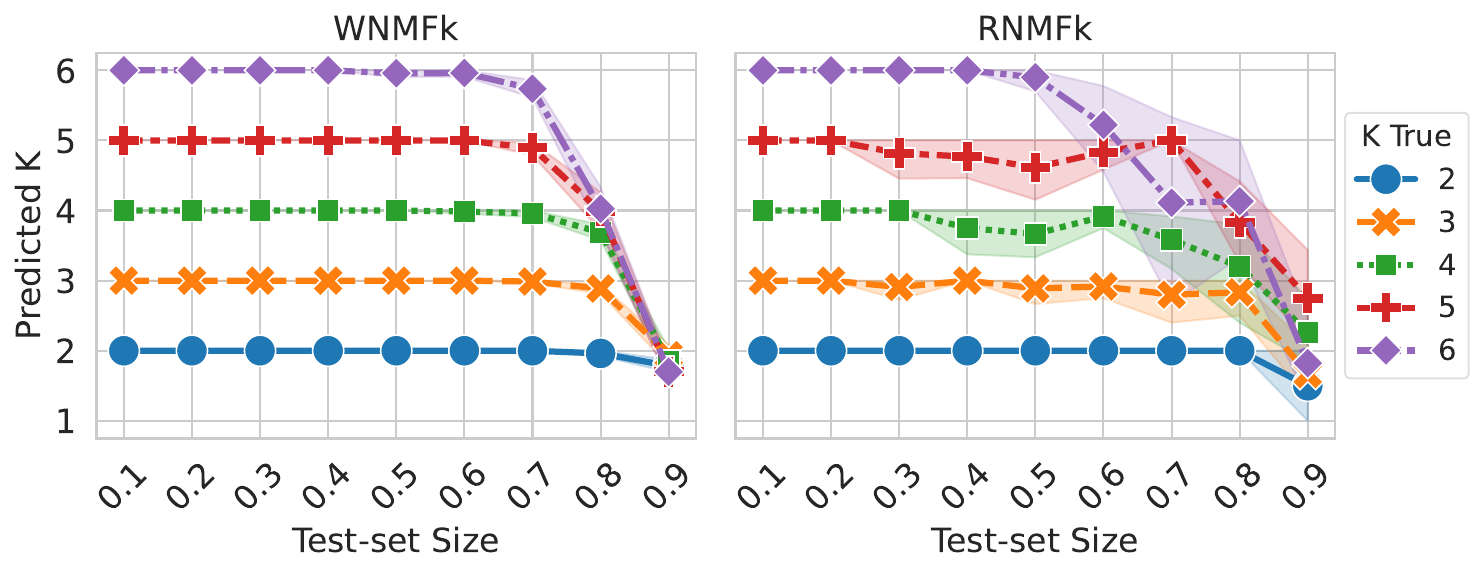} %
\caption{Results for $k$ predictions on Gaussian Data are shown for WNMFk and RNMFk methods across different data sparsity levels, represented by the increasing test-set size on the x-axis. The y-axis indicates the predicted $k$ values. The results demonstrate that $k$ predictions remain accurate for both methods until high levels of sparsity are reached, while WNMFk results in more accurate $k$ predictions. Line color and style differentiate between different true $k$ values. For instance, a solid line with circle markers represents decompositions where the matrix rank was $k=2$.
 \label{fig:ben_k_predictions}}
\end{figure}

Our final synthetic dataset, Gaussian Data, is analyzed in Figure \ref{fig:ben_k_predictions} for WNMFk and RNMFk, assessing their rank prediction at different test-set sizes. No Boolean settings are used. Color and line style differentiate true ranks; for example, an orange line with $x$ markers represents $k=3$. WNMFk predicts the correct rank until a test-set size of 0.7, after which predictions skew toward $k=1$. RNMFk follows a similar trend but loses rank prediction accuracy earlier at higher true ranks. For instance, at $k=6$ (purple line with diamond markers), lower rank predictions appear after a test-set size of 0.5.

\begin{figure}[t!]
\centering
\includegraphics[width=1.0\textwidth]{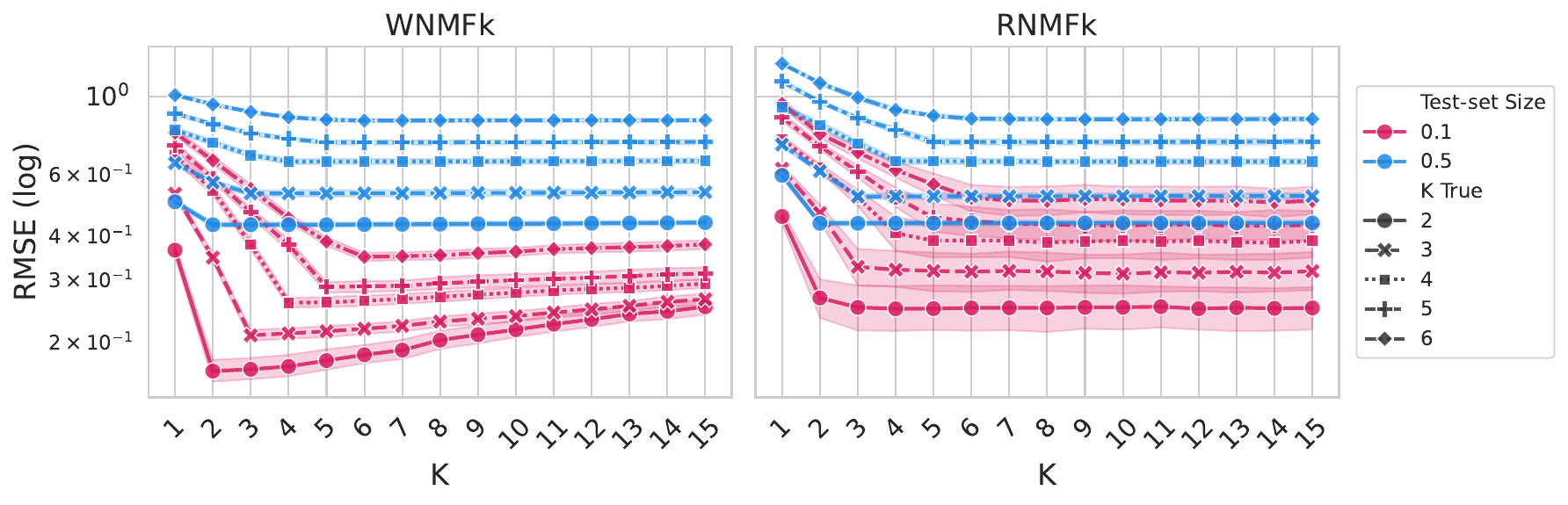} %
\caption{Link prediction performance on test-set sizes of 10\% (red lines) and 50\% (blue lines) is shown for WNMFk and RNMFk, with RMSE scores plotted on a log scale. The x-axis represents the $k$ values, displaying results at different $k$ decompositions. Line markers are used to differentiate between the true $k$ values. For example, solid red lines with circle markers represent matrices with a true rank of $k=2$ and a test-set size of 10\%.
\label{fig:ben_k_rmse}}
\end{figure}

We examine link prediction performance using RMSE in log scale for test-set sizes 0.1 and 0.5 in Figure \ref{fig:ben_k_rmse}, where blue lines represent 0.5 and red lines 0.1. Shapes differentiate true ranks; for example, square markers indicate $k=4$, with blue squares for $k=4$ at 0.5. For WNMFk, RMSE increases beyond the true rank, reflecting reduced performance, with the best results at the correct rank. This trend holds at higher sparsity (0.5 test-set size) but is less pronounced due to overall performance decline. RNMFk follows a similar pattern but shows more stable link prediction, as indicated by flatter RMSE lines, suggesting more consistent performance across ranks.

\begin{figure}[t!]
\centering
\includegraphics[width=1.0\textwidth]{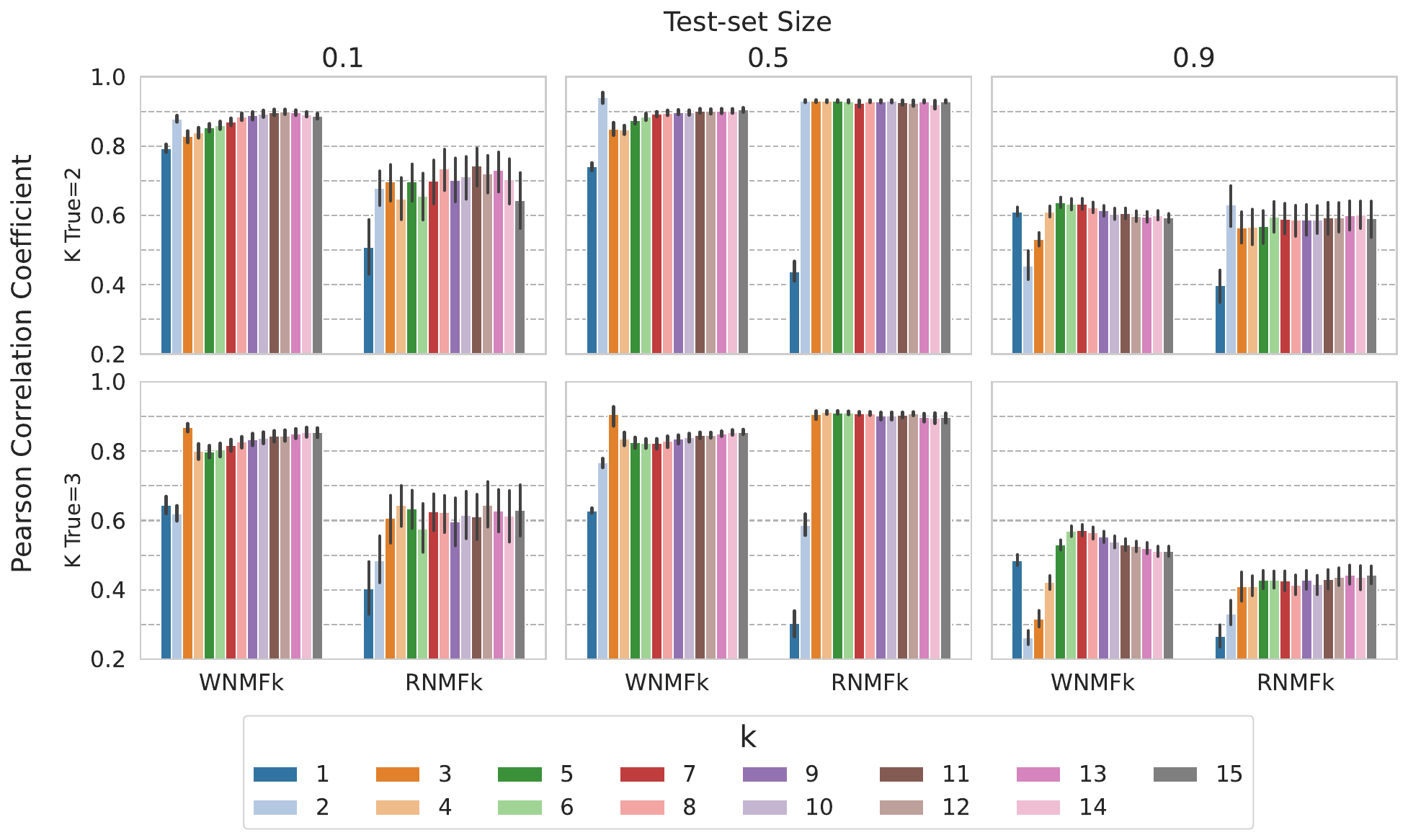} %
\caption{Pearson Correlation Coefficients with CI between UQ and errors at the test set are reported across different sparsity levels or test-set sizes that include 0.1, 0.5, and 0.9 (columns) and true matrix ranks $k$ including $k \in [2,3]$ (rows). The shared x-axis differentiates between the methods WNMFk and RNMFk using hues, while the shared y-axis represents the Pearson correlation values. Bars with different colors correspond to results at various $k$ decompositions. 
\label{fig:ben_uq_correlation}}
\end{figure}

For Gaussian Data, we analyze the correlation between UQ and errors in Figure \ref{fig:ben_uq_correlation} to assess whether models assign higher uncertainty to higher-error points. These trends align with the broader results in Appendix \ref{appendix-uq}, Figure \ref{fig:ben_uq_correlation_appendix}, which examines test-set sizes (0.1-0.9) and true ranks ($k=2$-$6$). Figure \ref{fig:ben_uq_correlation} shows Pearson Correlation Coefficients and CI between UQ and relative reconstruction errors (Equation \ref{eqn:rel_error}). Columns represent test-set sizes, rows correspond to true ranks $k=2$ and $k=3$, and the y-axis denotes correlation coefficients. Hue differentiates WNMFk and RNMFk, with color bars indicating decomposition results for each $k$.

Figure \ref{fig:ben_uq_correlation} shows that correlations between certainty and error remain high for RNMFk and WNMFk, confirming that UQ effectively assigns higher uncertainty to high-error points. Correlation is lower before the true rank and peaks at the correct rank, especially for WNMFk and smaller test-set sizes, indicating reliable certainty estimates. In the second row ($k=3$), the orange bars for WNMFk are notably higher, highlighting improved performance at this rank. Another key trend in Figure \ref{fig:ben_uq_correlation} is the reduction in correlation between certainty and error at the largest test-set size of 0.9. Our hypothesis for this behavior is that at higher sparsity levels, where the training set is significantly reduced, the uncertainty estimates may saturate at consistently high values across all test points, thereby reducing variability. We have also observed a relevant and similar result in Figure \ref{fig:dog_data_results_booleans_reject_option} where the fraction of abstained samples reduced at higher test set sizes.  When uncertainty does not vary meaningfully, its correlation with actual errors naturally diminishes. This saturation reflects a breakdown in the model's ability to provide meaningful uncertainty estimates, signaling that the model is operating in a regime of extreme sparsity. Under such conditions, there are significantly more missing links to predict while relying on minimal information from the training set. As a result, the model's uncertainty quantification methods become less reliable.

\begin{figure}[t!]
\centering
\includegraphics[width=0.7\textwidth]{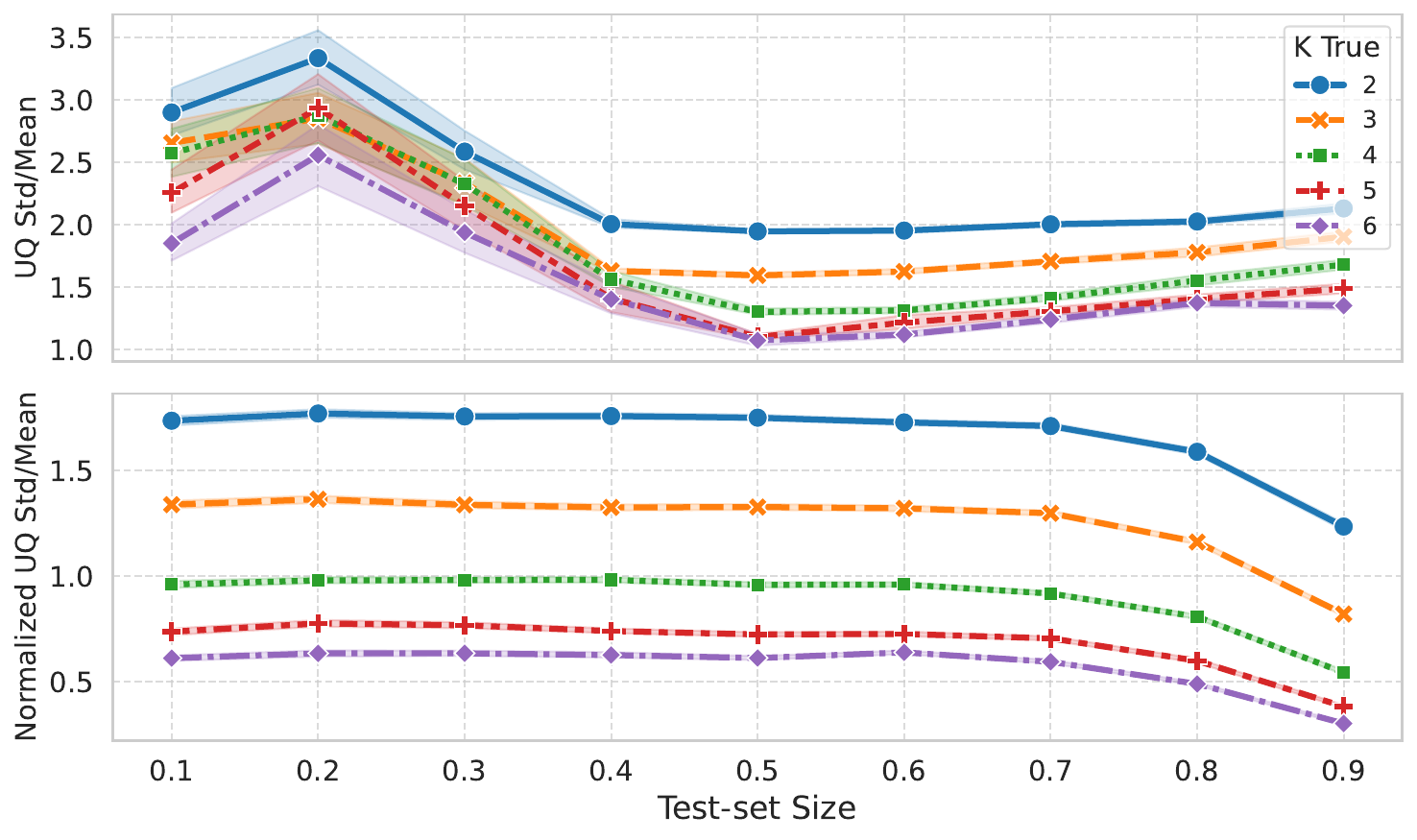} %
\caption{\textbf{Relative-Std-to-Mean Ratio of Uncertainty plot} provides a quantitative perspective on the behavior of uncertainty estimates across varying sparsity levels (or test set size) for each $k$ true. The top figure shows the standard deviation (std) at test points divided by the overall average of certainty (UQ). The bottom figure investigates certainty saturation by normalizing the std-to-mean uncertainty ratio at the top plot with the correlation between the UQ and errors at test points. \label{fig:uq_std_to_mean}}
\end{figure}

The "Relative-Std-to-Mean Ratio of Uncertainty" plot, shown in the top row of Figure \ref{fig:uq_std_to_mean}, provides a quantitative perspective on the behavior of uncertainty estimates across varying levels of sparsity (or test-set size). This metric is calculated for the test locations based on the uncertainty matrix $\mathbf{U}$ and is defined as:
\begin{align*} 
\text{SMR} = \frac{\sigma_{\mathbf{U}}}{\mu_{\mathbf{U}}}, \numberthis 
\end{align*}
where $\text{SMR}$ is the standard deviation (std)-to-mean ratio, $\sigma_{\mathbf{U}}$ is the standard deviation of UQ values at test locations, and $\mu_{\mathbf{U}}$ is the mean of UQ values at test locations, calculated as:
\begin{align*} 
\sigma_{\mathbf{U}} = \sqrt{\frac{1}{|\mathcal{I}_{\text{test}}|} \sum{(i,j) \in \mathcal{I}_{\text{test}}} (\mathbf{U}{ij} - \bar{\mathbf{U}})^2}, \quad \mu_{\mathbf{U}} = \frac{1}{|\mathcal{I}_{\text{test}}|} \sum{(i,j) \in \mathcal{I}_{\text{test}}} \mathbf{U}{ij}, \numberthis 
\end{align*}
where $\mathcal{I}_{\text{test}}$ is the set of test locations, and $\bar{\mathbf{U}}$ is the mean uncertainty value at the test locations.
This metric tracks the relationship between the variability of uncertainty (measured as the standard deviation of uncertainty) and the average uncertainty (mean of UQ) at the test points. By normalizing variability through its mean, the std-to-mean ratio captures whether the relative dispersion in uncertainty estimates decreases significantly as sparsity increases. In the top row of Figure \ref{fig:uq_std_to_mean}, the std-to-mean ratio drops until around a 0.5 test-set size and then stabilizes. Meanwhile, the correlation between errors and UQ remains high until a test-set size of 0.7 or 0.8, as shown in the Appendix Figure \ref{fig:ben_uq_correlation_appendix}. The drop in the std-to-mean ratio suggests that the variability of uncertainty at test locations is becoming more consistent relative to its mean but does not necessarily indicate saturation (since the correlation remains high). The normalized std-to-mean ratio plot, shown in the second row of Figure \ref{fig:uq_std_to_mean}, refines this measure by incorporating the correlation between UQ and errors. This metric is defined as:
\begin{align*} 
\text{NSMR} = \frac{\text{SMR} \cdot r}{\max(r)}, \numberthis 
\end{align*}
where $r$ is the vector of correlation values between UQ and reconstruction error. This normalization adjusts the standard deviation-to-mean ratio (SMR) by weighting it with the correlation between UQ and error, scaled by the maximum correlation value observed. This scaling emphasizes variability in uncertainty estimates that predict error rather than those that fluctuate without meaningful correlation to errors. By incorporating the correlation term, this metric provides a refined measure of how uncertainty dispersion (std-to-mean ratio) relates to actual predictive reliability. Their significance diminishes if uncertainty estimates are highly variable but poorly correlated with errors. Conversely, when variability in uncertainty remains well-correlated with errors, it suggests that the uncertainty estimates are informative.

In the second row of Figure \ref{fig:uq_std_to_mean}, higher values, observed until around a test-set size of 0.7, indicate that the uncertainty estimates remain variable and meaningfully related to errors. This suggests that UQ values are informative about prediction confidence. The observed decline beyond a test-set size of 0.8, where both the std-to-mean ratio in the first row of Figure \ref{fig:uq_std_to_mean} and the correlation in Figure \ref{fig:ben_uq_correlation} are low, suggests that uncertainty estimates are saturating. This saturation implies that the model's certainty values no longer distinguish between correct and incorrect predictions, reducing their usefulness at higher sparsity levels.

\subsection{PPI Data}
\label{subsec:ppi_results}

\begin{table}[t!]
\centering
\resizebox{\textwidth}{!}{

\begin{tabular}{lccccccccc}
\thickhline
\multirow{2}{*}{\textbf{Dataset}/Metric} & \multicolumn{9}{c}{Method} \\
\cmidrule(lr){2-10}
 & BNMFk & RNMFk & WNMFk & $\text{BNMFk}_{\textbf{lmf}}$ & $\text{RNMFk}_{\textbf{lmf}}$ & $\text{WNMFk}_{\textbf{lmf}}$ & LMF & symLMF \cite{ppi_symlmf} \\
\thickhline
\rowcolor{anti-flashwhite}
\multicolumn{10}{l}{\textbf{H.sapiens-extended}} \\
ROC AUC & $0.755 \pm 0.022$ & $0.941 \pm 0.001$ & $ 0.951 \pm 0.003$ & $\textbf{0.959} \pm 0.002$ & $0.955 \pm 0.001$ & $0.955 \pm 0.001$ & $0.949 \pm 0.001$ & $0.944 \pm 0.001$ \\
PR AUC & $0.884 \pm 0.009$ & $0.957 \pm 0.001$ & $ 0.964 \pm 0.003$ & $\textbf{0.969} \pm 0.002$ & $0.965 \pm 0.001$ & $0.965 \pm 0.001$ & $0.934 \pm 0.001$ & $0.955 \pm 0.001$ \\
UQ - ROC AUC & $\underline{0.762} \pm 0.027$ & $\underline{0.942} \pm 0.001$ & $\underline{0.954} \pm 0.003$ & $\underline{0.962} \pm 0.004$ & $\underline{0.956} \pm 0.001$ & $\underline{0.957} \pm 0.001$ & -- & -- \\
UQ - PR AUC & $0.880 \pm 0.010$ & $\underline{0.958} \pm 0.001$ & $\underline{0.966} \pm 0.002$ & $0.969 \pm 0.004$ & $\underline{0.966} \pm 0.001$ & $\underline{0.967} \pm 0.001$ & -- & -- \\
\midrule

\rowcolor{anti-flashwhite}
\multicolumn{10}{l}{\textbf{Brain}} \\
ROC AUC & $0.721 \pm 0.011$ & $0.943 \pm 0.001$ & $0.941 \pm 0.001$ & $\textbf{0.955} \pm 0.001$ & $0.954 \pm 0.001$ & $0.953 \pm 0.001$ & $0.931 \pm 0.001$ & $ 0.952 \pm 0.001$ \\
PR AUC & $0.849 \pm 0.004$ & $0.953 \pm 0.001$ & $0.948 \pm 0.001$ & $\textbf{0.962} \pm 0.001$ & $0.960 \pm 0.001$ & $0.959 \pm 0.002$ & $0.921 \pm 0.001$ & $ 0.957 \pm 0.001$ \\
UQ - ROC AUC & $\underline{0.727} \pm 0.010$ & $\underline{0.944} \pm 0.001$ & $\underline{0.945} \pm 0.002$ & $\underline{0.960} \pm 0.001$ & $\underline{0.955} \pm 0.001$ & $\underline{0.956} \pm 0.002$ & -- & -- \\
UQ - PR AUC & $0.840 \pm 0.004$ & $0.953 \pm 0.001$ & $\underline{0.952} \pm 0.002$ & $0.960 \pm 0.002$ & $\underline{0.961} \pm 0.001$ & $\underline{0.962} \pm 0.002$ & -- & -- \\
\midrule

\rowcolor{anti-flashwhite}
\multicolumn{10}{l}{\textbf{Disease of Metabolism}} \\
ROC AUC & $0.771 \pm 0.025$ & $0.958 \pm 0.006$ & $0.959 \pm 0.003$ & $\textbf{0.969} \pm 0.007$ & $0.966 \pm 0.008$ & $0.966 \pm 0.007$ & $ 0.968 \pm 0.002$ & $0.911 \pm 0.006$ \\
PR AUC & $0.861 \pm 0.014$ & $ 0.958 \pm 0.007$ & $0.957 \pm 0.003$ & $\textbf{0.968} \pm 0.008$ & $0.963 \pm 0.009$ & $0.963 \pm 0.008$ & $0.943 \pm 0.003$ & $0.926 \pm 0.005$ \\
UQ - ROC AUC & $\underline{0.785} \pm 0.024$ & $\underline{0.961} \pm 0.007$ & $\underline{0.962} \pm 0.003$ & $\underline{0.974} \pm 0.006$ & $\underline{0.968} \pm 0.008$ & $\underline{0.967} \pm 0.007$ & -- & -- \\
UQ - PR AUC & $\underline{0.863} \pm 0.017$ & $\underline{0.961} \pm 0.009$ & $0.957 \pm 0.003$ & $\underline{0.969} \pm 0.006$ & $\underline{0.965} \pm 0.009$ & $0.962 \pm 0.008$ & -- & -- \\
\midrule

\rowcolor{anti-flashwhite}
\multicolumn{10}{l}{\textbf{Liver}} \\
ROC AUC & $0.712 \pm 0.007$ & $0.944 \pm 0.001$ & $0.941 \pm 0.001$ & $\textbf{0.956} \pm 0.001$ & $0.954 \pm 0.001$ & $0.953 \pm 0.001$ & $0.933 \pm 0.001$ & $ 0.952 \pm 0.0004$ \\
PR AUC & $0.844 \pm 0.004$ & $0.954 \pm 0.001$ & $0.948 \pm 0.001$ & $\textbf{0.962} \pm 0.001$ & $0.961 \pm 0.001$ & $0.960 \pm 0.001$ & $0.925 \pm 0.001$ & $ 0.958 \pm 0.0004$ \\
UQ - ROC AUC & $\underline{0.719} \pm 0.009$ & $\underline{0.945} \pm 0.001$ & $\underline{0.945} \pm 0.002$ & $\underline{0.961} \pm 0.002$ & $\underline{0.955} \pm 0.001$ & $\underline{0.957} \pm 0.002$ & -- & -- \\
UQ - PR AUC & $0.833 \pm 0.003$ & $\underline{0.955} \pm 0.001$ & $\underline{0.951} \pm 0.002$ & $0.960 \pm 0.002$ & $\underline{0.962} \pm 0.001$ & $\underline{0.963} \pm 0.001$ & -- & -- \\
\midrule

\rowcolor{anti-flashwhite}
\multicolumn{10}{l}{\textbf{Neurodegenerative Disease}} \\
ROC AUC & $0.765 \pm 0.035$ & $0.968 \pm 0.006$ & $0.957 \pm 0.005$ & $\textbf{0.976} \pm 0.003$ & $0.973 \pm 0.002$ & $0.974 \pm 0.002$ & $ 0.974 \pm 0.002$ & $0.941 \pm 0.005$ \\
PR AUC & $0.873 \pm 0.015$ & $ 0.973 \pm 0.006$ & $0.961 \pm 0.005$ & $\textbf{0.981} \pm 0.003$ & $0.978 \pm 0.002$ & $0.979 \pm 0.002$ & $0.963 \pm 0.002$ & $0.952 \pm 0.004$ \\
UQ - ROC AUC & $\underline{0.769} \pm 0.029$ & $\underline{0.970} \pm 0.006$ & $\underline{0.959} \pm 0.005$ & $\underline{0.979} \pm 0.003$ & $\underline{0.975} \pm 0.002$ & $\underline{0.975} \pm 0.001$ & -- & -- \\
UQ - PR AUC & $0.859 \pm 0.016$ & $\underline{0.975} \pm 0.006$ & $\underline{0.962} \pm 0.006$ & $0.978 \pm 0.003$ & $\underline{0.979} \pm 0.002$ & $0.979 \pm 0.002$ & -- & -- \\
\thickhline
\end{tabular}

}
\caption{The performance of BNMFk, RNMFk, and WNMFk on five PPI datasets is reported and compared against their LMF-based extensions, indicated by the subscript \textbf{lmf}, i.e. $\text{WNMFk}_{\text{\textbf{lmf}}}$, $\text{BNMFk}_{\text{\textbf{lmf}}}$, and $\text{RNMFk}_{\text{\textbf{lmf}}}$. These results are also benchmarked against the LMF and symLMF methods. Average results are reported with $\pm$ two standard deviation. Performance metrics include ROC AUC and PR AUC, as well as ROC AUC and PR AUC with UQ-based weighting. Bolded scores indicate the best-performing methods for ROC AUC and PR AUC metrics across all approaches. For UQ-based scores, underlined values highlight instances where the inclusion of UQ-based weighting improved the scores compared to their non-UQ-weighted counterparts. Scores for symLMF are directly taken from \cite{ppi_symlmf}. Here, we run BNMFk with \textbf{kmeans} clustering for Boolean thresholding and RNMFk and WNMFk without Boolean thresholding. Reported LMF scores correspond to the LMF decomposition at various rank $k$ values that are predicted across the rank-$k$ estimations of BNMFk, RNMFk, and WNMFk for all cross-validation runs on the given PPI datasets.
}
\label{tab:metrics_LMF}
\end{table}
We next benchmark our methods against LMF and symLMF \cite{ppi_symlmf}. Table \ref{tab:metrics_LMF} presents the benchmark results on the PPI networks: \textit{H. sapiens}-extended, Brain, Disease of Metabolism, Liver, and Neurodegenerative Disease. For BNMFk, we employ k-means clustering for Boolean thresholding, as described in Section \ref{subsection:boolean_thresholding}. At the same time, RNMFk and WNMFk are evaluated without Boolean settings, following their original implementations as described in Sections \ref{subsection:wnmfk} and \ref{subsection:rnmfk}, respectively. Additionally, we include the LMF-extended results for $\text{WNMFk}_{\text{\textbf{lmf}}}$, $\text{BNMFk}_{\text{\textbf{lmf}}}$, and $\text{RNMFk}_{\text{\textbf{lmf}}}$, as described in Section \ref{subsection:lmf_ensemble}. Performance metrics include ROC AUC, PR AUC, and UQ-based scores, detailed in Section \ref{subsec:metrics}. 

Table \ref{tab:metrics_LMF} shows BNMFk has the lowest performance, while RNMFk and WNMFk consistently match or outperform LMF and symLMF across datasets. For example, on H.sapiens-extended, RNMFk achieves an ROC AUC of 0.941, compared to 0.949 for LMF and 0.944 for symLMF, with a higher PR AUC of 0.957 (vs. 0.934 for LMF and 0.955 for symLMF). Ensemble models yield higher performance, with $\text{BNMFk}_{\text{\textbf{lmf}}}$ achieving the highest scores across datasets, showing that combining models with LMF improves performance over standalone versions. On the Brain PPI dataset, BNMFk attains a PR AUC of 0.849, RNMFk 0.953, and WNMFk 0.948, while their LMF extensions reach 0.962, 0.960, and 0.959, respectively. These scores surpass LMF alone, indicating complementary model strengths under the ensemble framework. UQ-based weighting slightly improves scores in some cases, as shown by underlined ROC AUC and PR AUC values, though the gains are small for these datasets. These results on five real-world PPI networks validate the efficacy of our methods beyond synthetic datasets. The next section summarizes the technical capabilities and software development considerations behind our Python library.

\section{T-ELF: Python Library}
\label{sec:python_library}

We have publicly released all methods introduced in this paper through a user-friendly Python library, T-ELF \cite{TELF}\footnote{Available at \url{https://github.com/lanl/T-ELF}}. The library supports multi-processing via threading for concurrent operations, allowing each $k$ decomposition in the automatic model determination process to run in separate processes or GPUs. T-ELF uses Numpy \cite{harris2020array} for CPU operations and CuPy with CuPyx \cite{cupy_learningsys2017} for GPU-based dense and sparse matrix computations. HPC capabilities are enabled through OpenMPI, managed via \texttt{mpi4py} \cite{9439927, DALCIN20111124}, allowing further parallelization across HPC compute nodes for scalability. T-ELF is an installable Python package available via \texttt{pip} \cite{pypi} or Poetry, with example Jupyter Notebooks \cite{jupyter} and unit tests integrated into GitHub's CI pipeline. It includes comprehensive documentation, hosted via Sphinx and GitHub Pages \cite{sphinx_2020_sphinx}, with installation guides for environments like Conda \cite{anaconda}. To make the library easy to use, T-ELF follows an API style inspired by Scikit-learn \cite{pedregosa2011scikit}.

\section{Conclusion}
\label{sec:conclusion}

In this paper, we introduced novel matrix factorization extensions for link prediction: Weighted (WNMFk), Boolean (BNMFk), and Recommender (RNMFk), along with ensemble variants using logistic factorization. These methods address traditional limitations by integrating automatic model determination and UQ. Our framework heuristically estimates the optimal rank $k$ via a modified bootstrap-based stability and accuracy analysis, while UQ enhances reliability by abstaining from uncertain predictions. Compared with coordinate descent-based thresholding, we applied new Boolean thresholding techniques, including Otsu's method and k-means clustering for Boolean Matrix Factorization. Experiments on three synthetic datasets validated rank selection, assessed data sparsity effects, and tested link prediction performance in Boolean and non-Boolean settings. For the practical relevance, we benchmarked our methods against LMF and symLMF on five real-world PPI networks, demonstrating improved predictions. Finally, we released our methods as a user-friendly Python library on GitHub, supporting multi-processing, GPU acceleration, and HPC environments for large-scale applications.

\begin{acks}

This manuscript has been approved for unlimited release and has been assigned LA-UR-25-22115. This work was funded by a grant HDTRA1242032(CB11198) of BSA, from the Defense Threat Reduction Agency (DTRA) of the U.S. Department of Defense (DoD). The funders had no role in study design, data collection and interpretation, or the decision to submit the work for publication. Funds for the demonstration and/or assessment work were provided by the Los Alamos National Laboratory Technology Evaluation \& Demonstration program. The research was also supported by LANL Institutional Computing Program, and  by the U.S. DOE NNSA under Contract No. 89233218CNA000001.

\end{acks}

\bibliographystyle{ACM-Reference-Format}
\bibliography{main}

\appendix
\section{Appendix}
\label{appendix_sec}

In the Appendix, we provide additional results presented in Figures \ref{fig:dog_data_results}, \ref{fig:dog_data_results_booleans}, \ref{fig:dog_data_results_booleans_reject_option}, and \ref{fig:ben_uq_correlation}.  With these figures, a subset of these results was presented in the main text for simplicity and clarity. We include the remaining details in the Appendix with Figures \ref{fig:dog_data_results_appendix} and \ref{fig:ben_uq_correlation_appendix} to view the results comprehensively.

\subsection{Dog Data Results (Expanded Version)}
\label{appendix-dog}

\begin{figure}[t!]
\centering
\includegraphics[width=1.0\textwidth]{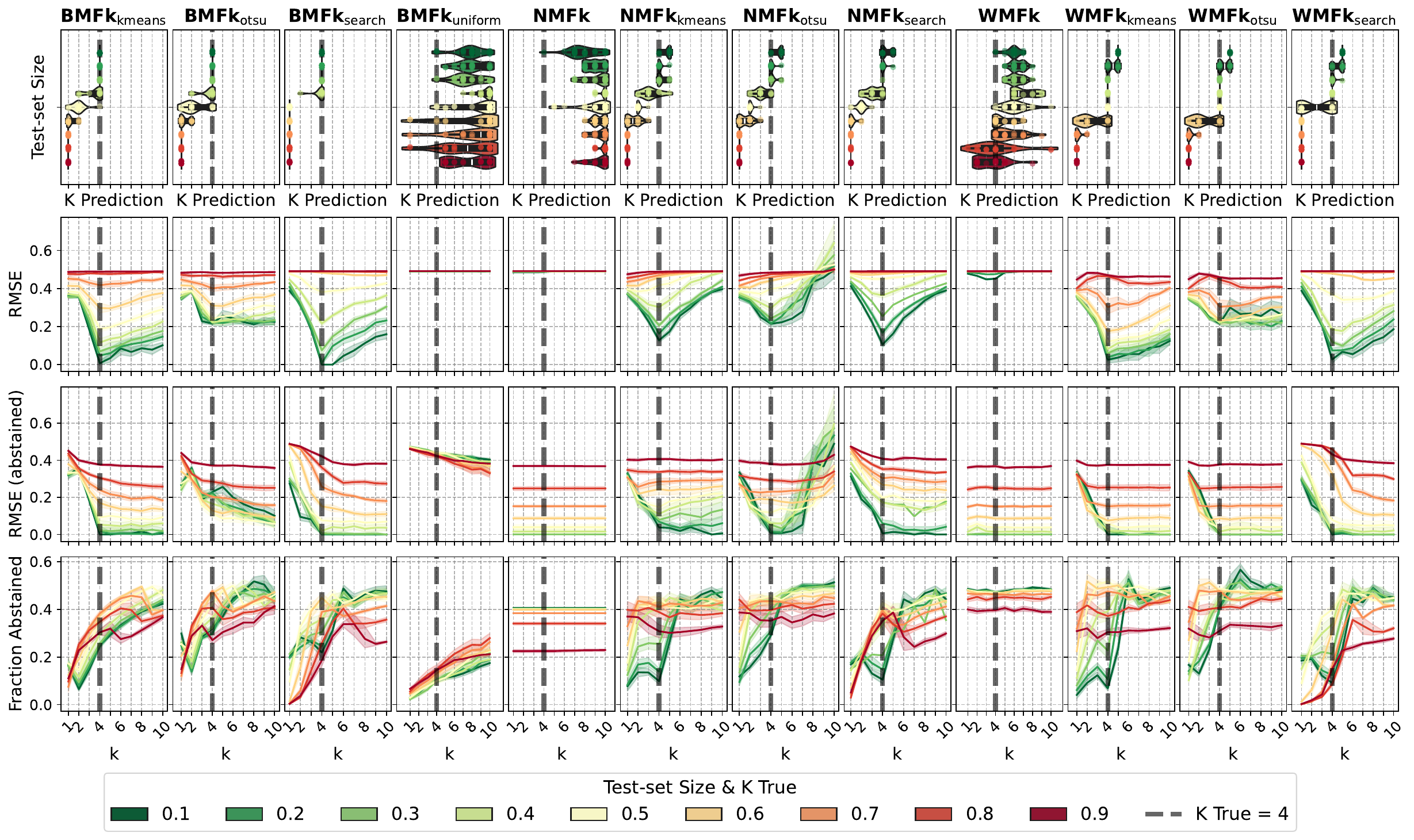} %
\caption{Results for Dog Data across each method, including Boolean and non-Boolean variations (columns). The Boolean thresholding techniques are denoted with the subscripts of \textbf{kmeans}, \textbf{otsu}, and \textbf{search} (coordinate descent). For example, the absence of a subscript, WNMFk, refers to not using a Boolean thresholding technique with that given method. For BNMFk, sub-script \textbf{uniform} refers to running BNMFk without the factor Boolean thresholding. The first row presents violin plots visualizing the rank $k$ predictions at different sparsity or test-set sizes. The second row displays RMSE scores for the missing link prediction performance test set. The third row shows RMSE scores calculated for non-abstained predictions when applying UQ. The fourth row illustrates the fraction of abstained predictions, indicating the proportion of cases where the model chose to abstain ("I do not know") rather than make a prediction. The results are reported for each rank $k$, the x-axis in the plots, while the dark and dashed vertical line across each column is the true rank $k=4$.
 \label{fig:dog_data_results_appendix}}
\end{figure}

Our results on the Dog Data are presented in Figure \ref{fig:dog_data_results_appendix} and provide a more detailed view of method performance under both Boolean and non-Boolean settings. This figure expands upon the results shown in Figures \ref{fig:dog_data_results}, \ref{fig:dog_data_results_booleans}, and \ref{fig:dog_data_results_booleans_reject_option} by including all methods together and also showing results for RMSE, RMSE abstained, and Fraction Abstained.

The first row presents violin plots of predicted ranks $k$ across test-set sizes, illustrating how each method estimates the correct rank under varying sparsity levels. The second row reports test-set RMSE, highlighting link prediction accuracy. The third row examines RMSE after UQ, considering only non-abstained predictions. The fourth row shows the fraction of abstained predictions, where higher values indicate a lower coverage rate. The trends in Figure \ref{fig:dog_data_results_appendix} align with the main text. BNMFk, using boolean threshold techniques from Section \ref{subsection:boolean_thresholding} (denoted by subscripts), predicts $k=4$ more frequently than NMFk and WNMFk, especially at lower sparsity. BNMFk's rank predictions shift downward as test-set size grows, reflecting adaptation to increased sparsity. Non-Boolean methods (WNMFk, NMFk, and $\text{BNMFk}_{\textbf{uniform}}$) often predict higher ranks, capturing different structural properties. RMSE results confirm that Boolean thresholding improves link prediction, particularly for NMFk and WNMFk. Performance declines with larger test sets, as indicated by the color shift from green to red, but Boolean methods maintain the lowest RMSE near the correct rank. RMSE changes from abstention (third row) show that UQ reduces errors by ignoring uncertain predictions, leading to lower RMSE scores than in the second row. The fraction of abstained predictions (fourth row) increases with test-set size, yet confidence improves near the correct rank for several of the methods. This figure adds granularity beyond the main text, reinforcing conclusions on rank prediction accuracy, RMSE trends, and UQ's role in improving predictions.

\subsection{Uncertainty-Error Correlation Results (Expanded Version)}
\label{appendix-uq}

\begin{figure}[t!]
\centering
\includegraphics[width=1.0\textwidth]{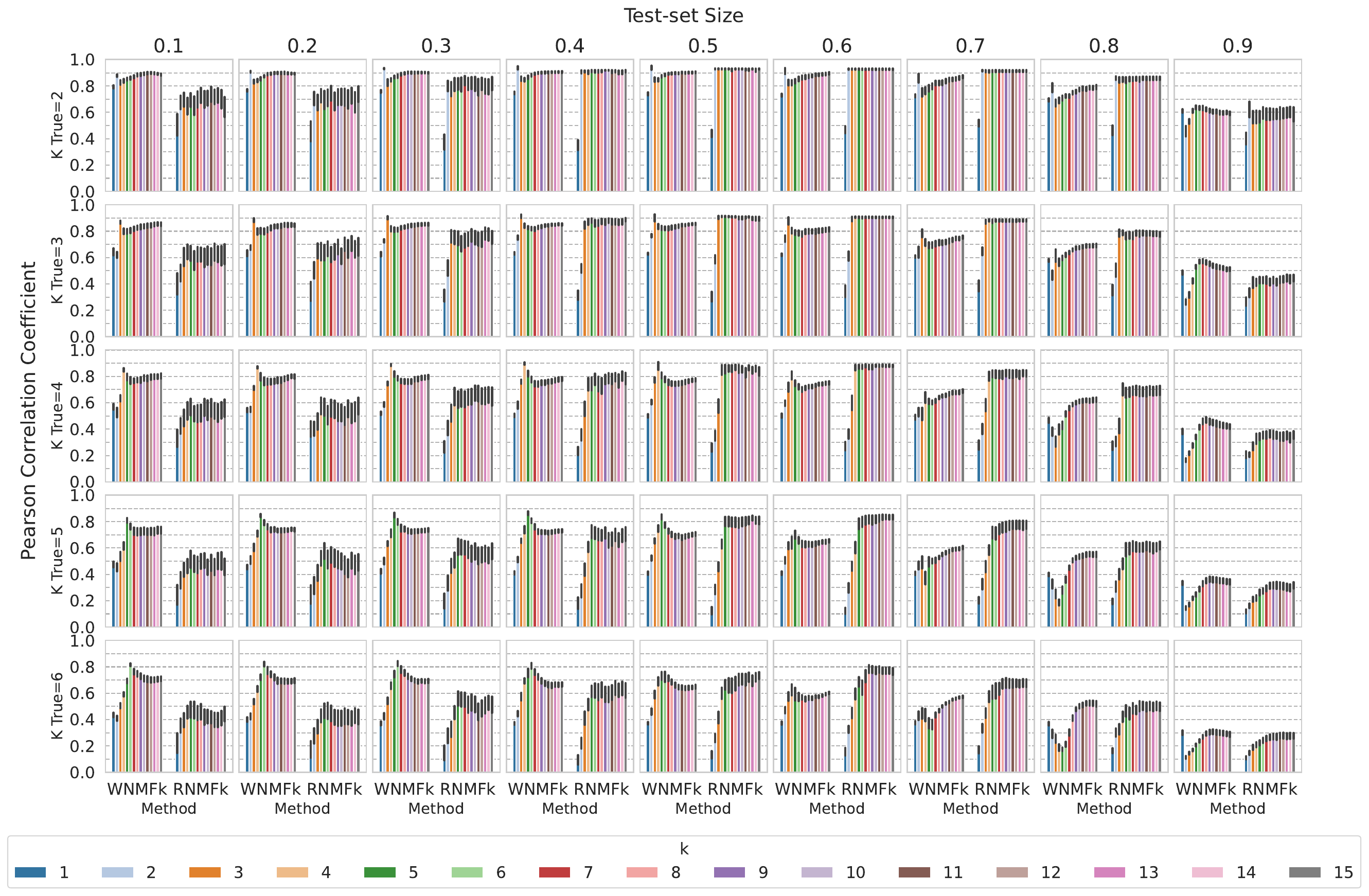} %
\caption{Pearson Correlation Coefficients with CI between UQ and errors are reported across test-set sizes (columns) and true ranks $k$ (rows). The x-axis differentiates WNMFk and RNMFk by hue, while the y-axis represents correlation values. Colored bars indicate results for different $k$ decompositions.
\label{fig:ben_uq_correlation_appendix}}
\end{figure}

Figure \ref{fig:ben_uq_correlation_appendix} presents a more detailed analysis of the correlation between UQ and prediction errors in the Gaussian Data. This figure extends the results shown in Figure \ref{fig:ben_uq_correlation} by covering a broader range of test-set sizes (0.1 to 0.9, in increments of 0.1) and true ranks ($k$ from 2 to 6).

The Pearson Correlation Coefficients and CIs are reported across different sparsity levels (columns) and true matrix ranks (rows). The x-axis differentiates WNMFk and RNMFk by hue, while the y-axis represents Pearson correlation values, with each color bar corresponding to a specific decomposition rank $k$. Here, WNMFk and RNMFk do not use Boolean thresholding. The observed trends align with those in the main text across a broader range of test-set sizes and true ranks. Correlation between uncertainty and error is generally high, as models assign greater uncertainty to locations with larger prediction errors. Correlation peaks at the true rank, especially for lower test-set sizes in WNMFk, but declines at extreme sparsity levels (test-set sizes above 0.7 or 0.8). This supports the hypothesis that uncertainty saturates at high values across all test points, reducing variability and weakening its correlation with actual errors. The figure provides a more comprehensive view of UQ and error interactions, complementing the main text results.

\end{document}